\documentclass[11pt]{article}

\usepackage[margin=1in]{geometry}
\usepackage[T1]{fontenc}
\usepackage[utf8]{inputenc}
\usepackage{lmodern}   
\usepackage{microtype}
\usepackage{amsmath,amssymb}
\usepackage{booktabs}
\usepackage{float}
\usepackage{tabularx}
\usepackage{multirow}
\usepackage{graphicx}
\usepackage{caption}
\usepackage{xcolor}
\usepackage[round]{natbib}
\usepackage[colorlinks=true,linkcolor=blue!60!black,citecolor=blue!60!black,urlcolor=blue!60!black]{hyperref}

\newcommand{\nfm}{NFM}
\newcommand{\nfmmh}{NFM-MH}
\newcommand{\st}{ST}
\newcommand{\sttwo}{ST-2H}
\newcommand{\stl}{ST-L}
\newcommand{\xhat}{\hat{x}}
\newcommand{\xs}{x^{*}}

\newcommand{\nIter}{15}
\newcommand{\trainEpochs}{420{,}000}

\newcommand{\paramsNFMexact}{1{,}325{,}720}
\newcommand{\paramsNFMMHexact}{1{,}326{,}695}
\newcommand{\paramsUpdater}{843{,}214}
\newcommand{\paramsSTexact}{1{,}328{,}102}
\newcommand{\paramsSTTWOexact}{1{,}331{,}537}
\newcommand{\nScored}{5{,}200}
\newcommand{\nHard}{1{,}040}
\newcommand{\nTies}{1{,}200}
\newcommand{\nUnseen}{880}
\newcommand{\nMonitored}{400}
\newcommand{\nHeavy}{975}
\newcommand{\paramsMHexact}{1{,}331{,}079}
\newcommand{\paramsSTearly}{1{,}327{,}418}
\newcommand{\paramsMHearly}{1{,}330{,}395}
\newcommand{\paramsSTLexact}{1{,}332{,}422}

\title{One-Shot Localisation of the Global Minimum\\ of a Noisy One-Dimensional Function:\\ A Controlled Comparison of an Iterative Neural Minimizer,\\ Set Transformers and Classical Estimators}

\author{Qusay Muzaffar$^{1}$ \and David Levin$^{2}$ \and Michael Werman$^{1}$\\[4pt]
{\small $^{1}$Department of Computer Science, The Hebrew University of Jerusalem, Jerusalem, Israel}\\
{\small $^{2}$Department of Applied Mathematics, Tel Aviv University, Tel Aviv, Israel}\\
{\small\texttt{qusay.muzaffar@mail.huji.ac.il}, \texttt{levin@tauex.tau.ac.il}, \texttt{michael.werman@mail.huji.ac.il}}}
\date{}

\ifdefined\pdfinfoomitdate\pdfinfoomitdate=1\pdftrailerid{}\fi
\begin{document}
\maketitle

\begin{abstract}
We study passive global optimisation: from twenty noisy samples of an unknown one-dimensional function, with no further queries, estimate where its global minimum lies. We build a benchmark and a self-checking harness and use them for a controlled comparison. The Neural Function Minimizer (\nfm), which we introduce, moves a position across the domain for fifteen steps and re-reads the samples from it at every step; a same-size Set Transformer answers in one pass. In a $2\times2$ design each has one output head or an added mixture head, with the same data and inputs and three seeds each. Within the training families the two are tied on location error, and the mixture loss lowers it by about 0.45 points of the domain in both. The \nfm's answers have lower regret, by 2.90 and 3.12 points of the range with one and two heads (significant after correction, and again on a block of fresh cases), mainly because its right-basin answers lie closer to the floor; a Set Transformer looped to the \nfm's computation does not reproduce this within the training families, though it has the lower error on unseen functions (exploratory). The \nfm{} also differs from the Set Transformers in encoder, local read, activation, losses and training time. At near-tied minima the form of the answer decides: point answers hedge in about a quarter of exact ties, mixture modes almost never, mixture means in more than 80\%. Gaussian-process estimators, with no training, have the lowest regret and are the most accurate on unseen functions. We will release the benchmark, the models and the harness.
\end{abstract}

\section{Introduction}

Global optimisation of an expensive black-box function usually proceeds by querying: a surrogate is fitted to what has been observed, an acquisition rule proposes the next point, and the loop repeats \citep{shahriari2016taking, snoek2012practical}. This paper studies the situation in which no further query is possible. A fixed set of $n=20$ noisy samples of an unknown function $f:[0,1]\to\mathbb{R}$ is all there is, and the task is to name the location $\xs$ of its global minimum. The situation arises when a batch of evaluations has already been made, a sweep of one parameter of an experiment, a simulation or a training run, and a setting must be chosen from it without further runs; it is also the zero-query limit of Bayesian optimisation, which isolates how much an estimator extracts from the data it is given. The problem is one of inference rather than search: the information about $\xs$ is already in the samples, and the question is how much of it an estimator can extract. In statistics it is the estimation of a minimiser under passive design \citep{muller1989adaptive, akhavan2024estimating}.

The classical answer is to fit a smooth model and take its argmin. A learned answer is to train a network, on a large synthetic family of functions, to map the sample set directly to $\xs$ \citep{amos2023tutorial, chen2022learning}. Such a network can compute its answer in different ways, and this paper asks whether one particular way, iteration, changes what the estimator does. Our model, the Neural Function Minimizer (\nfm), is iterative: an encoder summarises the sample set once, and an iterator then moves a position $x_t$ across the domain for $T=\nIter$ steps; at every step the model reads the samples near $x_t$ through a learned kernel, attends to all twenty samples from where it now stands, and revises its state. Its single-pass counterpart is a Set Transformer \citep{lee2019set} of the same size, trained on the same data and given the \nfm's inputs, which maps the set to an answer in one pass (\st).

A comparison of two architectures is only as clean as what is held fixed. We therefore compare them in a $2\times2$ design. The first factor is the computation: the walk of the \nfm{} or the single pass of the \st. The second is the output: one head that gives a point and a spread, or a second head, added to the same network, that gives a five-component Laplace mixture over the location of the minimum, in the manner of a mixture density network \citep{bishop1994mixture}, trained with the mixture likelihood alongside the original loss. The two two-head models, \nfmmh{} and \sttwo, carry the same mixture head with the same mixture loss (their point and spread losses differ as the base models' do, Section~\ref{sec:baselines}), and all four models have the same size, the same training data, the same inputs, the same recipe and three training seeds each. A two-head model answers in three ways from one network: with its point head, with the location of its heaviest mixture component (the \emph{mode}), or with the weighted mean of its components (the \emph{mixture mean}). Classical estimators, the Gaussian-process (GP) posterior argmin \citep{rasmussen2006gaussian} and the GP's posterior distribution over the minimiser \citep{hennig2012entropy} in particular, answer the same cases.

The design separates three effects. We call the first the effect of the walk, for short: it is the effect of the \nfm{} as a whole against the Set Transformer, and the \nfm{} differs in more than the walk (below). It is about the same with one head and with two: it does not change the mean location error, on which the two architectures are tied with either head, but it lowers regret (the excess function value at the answer, in percent of the function's range) by about three points, mostly through answers in the right basin, which lie closer to the floor, and its own uncertainty ranks its errors better. The effect of the mixture loss is about the same in both designs: it lowers the mean location error by about 0.45 points of the domain and regret by 2.5 to 2.7 points, whatever computes the answer. And the effect of the form of the answer, read from one network, is the largest of the three on how often an answer hedges where two valleys are nearly equally deep: the point heads hedge between the valleys in about a quarter of exact ties (24 to 29\%), the modes commit, and the mixture means hedge in more than 80\%. On the constructed ties we detect no effect of the architecture on how often the point answer hedges; the form of the answer decides.

Our contributions are the following.

\begin{enumerate}
\item A benchmark for passive argmin estimation with an explicit noise model, $\sigma=\lambda d$ with $d$ the depth of the global well, and an evaluation harness that scores every estimator on identical cases: in-family cases at four scored and two hard noise levels, constructed ties, heavy-tailed noise, and two out-of-family sets; tie-aware measures and regret alongside location error; a full set of uncertainty measures, with the GP's distribution over the minimiser as the classical reference; three training seeds for every learned model, statistics that resample seeds and cases, and declared families of comparisons with a correction for multiple testing. The harness checks itself and fingerprints its cases and checkpoints; two runs on different machines use identical cases and agree to rounding (Section~\ref{sec:benchmark} and Appendix~\ref{app:harness}).
\item A controlled comparison of iterative and single-pass computation in a $2\times2$ design, with a looped Set Transformer of matched computation as a further control (Sections~\ref{sec:results-accuracy}, \ref{sec:results-walk}, \ref{sec:results-calibration} and~\ref{sec:results-looped}).
\item An observation about ties within one network: point answers trained on distance hedge between nearly equal valleys, even when noiseless samples resolve the tie, while the modes of the same network commit and its mixture means hedge most; the pattern is consistent with the loss, with averaging and with smoothing (Section~\ref{sec:results-tie}).
\item The \nfm{} architecture and its training: an iterative model whose update module reads the samples by attention at every step, with inputs that separate the shape of the samples from their scale (which we give to the Set Transformers as well), and the design choices and monitoring that keep its update module in use, including a cheap check that detected a large module that had silently stopped working (Section~\ref{sec:model}).
\end{enumerate}

The \nfm{} differs from the Set Transformers in more than the walk: it also has a different encoder, a local read, a learned activation and loss terms on its trajectory, and trains about ten times as long as the single pass. Our experiments therefore attribute the regret effect to the \nfm{} as designed and trained; a looped Set Transformer of matched computation narrows this down without settling it. The order in which results were seen is stated in Section~\ref{sec:results-selection}, and what the evidence does not show is collected under Limitations in Section~\ref{sec:discussion}.

\section{Related work}

\paragraph{Nonparametric estimation of the location of an extremum.} Estimating where a regression function attains its extremum from noisy samples is a classical problem in statistics, usually addressed by smoothing the data and locating the extremum of the estimate, with bandwidths chosen for the location rather than for the curve \citep{muller1989adaptive}; minimax rates for the minimiser and the minimum value under passive design are given by \citet{akhavan2024estimating}. Local polynomial fits with moving weights, whose approximation power is analysed by \citet{levin1998approximation} for moving least squares, are the basic tool of this approach \citep{fan1996local}. Our problem is this one, at a fixed small sample size and with multimodal functions; the spline, local-quadratic and GP argmin baselines are estimators of this kind, and our learned models replace the smoother by a map trained on a family of functions.

\paragraph{Bayesian and surrogate-based optimisation.} Gaussian-process surrogates with acquisition rules such as expected improvement or lower confidence bound \citep{srinivas2010gaussian, snoek2012practical, shahriari2016taking} are the standard tools for expensive black-box minimisation. They assume the ability to query. The zero-query limit of this family is the argmin of the posterior mean fitted to the given samples, which is one of our baselines; entropy search \citep{hennig2012entropy} and predictive entropy search \citep{hernandez2014predictive} work with the posterior distribution of the argmin itself, whose median and mode are two further zero-query estimators that we evaluate, and whose spread is the classical counterpart of a learned model's uncertainty.

\paragraph{Learning to optimise and amortised optimisation.} Recurrent optimisers trained to produce update steps \citep{andrychowicz2016learning, chen2017learning} and, more generally, amortised optimisation \citep{amos2023tutorial, chen2022learning} replace hand-designed iteration with a learned one. Closest in structure is algorithm unrolling \citep{gregor2010learning, monga2021algorithm}, in which the steps of an iterative algorithm become the layers of a network with learned parameters; the \nfm{} is an unrolled search of this kind. It is an amortised estimator with an iterative inner structure, but it is trained to locate the minimum of a function it never queries; its iteration re-reads a fixed sample set rather than acquiring new evaluations.

\paragraph{Recurrent glimpses and iterative refinement.} The \nfm's loop of moving a position, reading the data near it and revising a state has close relatives outside optimisation. The recurrent attention model of \citet{mnih2014recurrent} moves a location across an image, reads a small glimpse around it, updates a recurrent state and chooses where to look next; its glimpse is hard and trained by reinforcement learning, whereas the \nfm's Cauchy window is soft and differentiable. DRAW \citep{gregor2015draw} reads an image through a differentiable window with a learned centre and width, a close relative of our local read. RAFT \citep{teed2020raft} estimates optical flow by keeping a current estimate, looking up features around it, updating a recurrent state and emitting an increment, with the same weights at every iteration, which is the loop of our local read, iterator and updater. The updater, whose state cross-attends to the sample tokens at every step with the same weights, belongs to the family of Slot Attention \citep{locatello2020object} and the Perceiver \citep{jaegle2021perceiver}, and its depth comes from repeating one weight-tied block, as in Universal Transformers \citep{dehghani2019universal}. We know of no model in these lines that walks over a fixed set of samples of a function it cannot query in order to locate its minimum; the parts of the \nfm{} have relatives, and what is new is their combination for this task.

\paragraph{Set-structured inputs and in-context inference.} Deep Sets \citep{zaheer2017deep} and the Set Transformer \citep{lee2019set} give permutation-invariant maps from a set of observations to an output, built on attention \citep{vaswani2017attention}; neural processes \citep{garnelo2018neural}, with attention in attentive neural processes \citep{kim2019attentive}, and prior-data-fitted transformers \citep{muller2022transformers}, trained like our models on draws from a simulated prior and applied to tabular data by TabPFN \citep{hollmann2023tabpfn}, learn to perform Bayesian-style inference in a single forward pass over a context set, transformer neural processes \citep{nguyen2022transformer} do so with attention, prior-data-fitted networks have been applied to Bayesian optimisation \citep{muller2023pfns4bo}, and transformers have been trained to act as hyperparameter optimisers \citep{chen2022towards}. Mixture density networks \citep{bishop1994mixture} let such a model output several modes instead of one point. Our Set Transformers represent the single-pass family, with one output head and with a point head and a mixture head.

\paragraph{Looped and weight-tied models.} Applying one block repeatedly with shared weights was proposed for sequence models by the Universal Transformer \citep{dehghani2019universal}, and has been studied as a way to learn algorithms that extend to harder instances \citep{schwarzschild2021can, bansal2022end} and as a model of in-context learning \citep{giannou2023looped, yang2024looped}, in which transformers are trained, as ours are, on draws from simple function classes \citep{garg2022what}. Iterative amortised inference \citep{marino2018iterative} refines an approximate posterior by repeated learned updates instead of one encoder pass, the counterpart of the walk against a single pass in variational inference. Our looped Set Transformer (Section~\ref{sec:results-looped}) is a control of this kind; its loss supervises only the last step, whereas the walk's trajectory term supervises every step, a form of deep supervision \citep{lee2015deeply}.

\paragraph{Adaptive computation, uncertainty and evaluation.} Adaptive computation time \citep{graves2016adaptive}, PonderNet \citep{banino2021pondernet} and deep equilibrium models \citep{bai2019deep} vary how much a recurrent model iterates. For the evaluation of predictive uncertainty we follow proper scoring rules and the calibration--sharpness view \citep{gneiting2007strictly}, quantile calibration \citep{kuleshov2018accurate}, calibration by spread level \citep{levi2022evaluating}, and risk--coverage curves \citep{geifman2019bias}; for statistics over few training runs and several data sets, \citet{demsar2006statistical}, \citet{holm1979simple}, \citet{bouthillier2021accounting} and \citet{agarwal2021deep}.

\section{Problem and benchmark}
\label{sec:benchmark}

\subsection{Task}

A case is a function $f$ on $[0,1]$ with a unique global minimiser $\xs$, a set of $n=20$ sample locations $x_1,\dots,x_n$, and noisy observations $y_i=f(x_i)+\varepsilon_i$. An estimator receives $\{(x_i,y_i)\}$ and returns $\xhat\in[0,1]$. The primary score is the location error $|\xhat-\xs|$, reported in percent of the domain. The learned models receive, in addition to the samples, quantities computed from the samples alone by a cubic smoothing spline fitted to them: the spline's slope $y_i'$ at each sample location, the spline's B-spline coefficient vector $c$ (padded to length 20), a starting point for the rollout (the spline's argmin), a resolution radius $\delta$, and five global descriptors $\mathbf{g}$ (an estimated noise scale, the resolution radius, a well-separability score, a whiteness score and the scale of the samples). All four learned models receive these inputs in the same scale-free form, built by the same code (Section~\ref{sec:model}), except the starting point, which only the walk models use; their answer depends little on it (Appendix~\ref{sec:results-path}). None of these uses the true function. The classical estimators receive the samples $\{(x_i,y_i)\}$ and compute whatever they need from them. Appendix~\ref{app:bench} gives every formula.

\subsection{Function families and noise}

\paragraph{Families.} Training and in-family evaluation functions are drawn from thirteen parametric families on $[0,1]$, with the mixture weights in parentheses: bowls $s(x-m)^p$ with $p\in\{2,4,6,8\}$ (0.09); random polynomials of degree 3 to 8 (0.08); sums of two to four sinusoids, half of them chirped, on a quadratic bowl (0.09); a regular cosine lattice on a quadratic bowl (0.09); tilted quartic double wells (0.10); two to four Gaussian wells on a shallow bowl (0.08); random cubic B-splines (0.06); deceptive functions, a deep narrow well beside a broad bowl whose floor lies elsewhere (0.06); cosine ripples whose amplitude grows (0.09) or decays (0.06) across the domain; a cosine lattice under an exponential envelope (0.06); plateaus with narrow valleys (0.06); and spectral functions (0.08), $f(x)=b(x-m)^2+\tau x+A\sum_{k=1}^{K}c_k\cos(2\pi kx+\phi_k)$ with $c_k\propto k^{-\xi/2}$, $\xi\sim U(1,3)$ and $K\sim U\{8,\dots,24\}$, whose power spectrum falls as $k^{-\xi}$. Here $m\sim U(0.05,0.95)$ is a centre drawn for every function. In every family the global minimiser $\xs$ is located numerically on a 3{,}000-point grid, and a draw is rejected and redrawn unless $\xs\in[0.05,0.95]$, the range of $f$ lies in $[10^{-4},10^{6}]$, the well depth $d$ defined below is positive, and the global basin is not too narrow. Appendix~\ref{app:families} gives every formula, every parameter distribution and the acceptance tests, and Table~\ref{tab:famstats} the measured properties of each family. Function values are shifted so that the minimum over the twenty clean samples is zero and are not rescaled; the noise is added after the shift.

\paragraph{Sampling.} The $n=20$ locations come from one of three samplers, chosen with probabilities 0.50, 0.25 and 0.25: uniform on $[0.001,0.999]$; Gaussian, with centre drawn from $U(0.1,0.9)$ and standard deviation from $U(0.10,0.30)$, clipped to $[0.001,0.999]$; and Beta$(a,b)$ with $a,b\sim U(0.5,3)$. The points are sorted and pushed apart to a minimum spacing of $0.005$. A draw is rejected, and sampler and points are drawn again, unless the twenty points are distinct and at least one lies $0.02$ or more to each side of the global minimum, so that every case is answerable in principle. Because the Gaussian and Beta samplers miss the minimum more often, the mixture that survives rejection is about 58/20/22 rather than 50/25/25 (3{,}025, 1{,}057 and 1{,}118 of the \nScored{} in-family evaluation cases), in training and in evaluation alike. The full algorithm is in Appendix~\ref{app:sampler}.

\paragraph{Noise.} The noise is independent, zero-mean Gaussian, and it is scaled to the decision rather than to the function. Let $d$ be the depth of the global well: the difference between the lowest rival local minimum and the global minimum of $f$ on the 3{,}000-point grid, where an endpoint towards which the curve descends counts as a rival; when there is no rival at all, $d$ is the drop from the lower of the two endpoints to the global minimum. Each observation receives an independent draw $\varepsilon_i\sim\mathcal N(0,\sigma^2)$ with $\sigma=\lambda d$, with $\lambda\in\{0,0.25,0.5,1,2,4\}$ drawn with equal probability in training and fixed per cell in evaluation; $\lambda=0$ cases are noiseless. Scaling to $d$ is what makes a noise level mean the same thing on every family. Scaled to the range of $f$ or to its standard deviation instead, the same $\lambda$ would give very different difficulties: over three independent sets of 80 draws per family, the median ratio of range to $d$ differs between the most and the least extreme family (double wells and deceptive functions) by a factor of 130 to 240, and the ratio of standard deviation to $d$ by a factor of 120 to 230. At $\lambda=1$ the noise equals the feature that distinguishes the answer from its nearest competitor. Even if all twenty samples could be spent comparing the two wells, the signal-to-noise ratio of that comparison would be $d/(\sigma\sqrt{4/n})=\sqrt{5}/\lambda$, with ten samples on each well, which is 2.2 at $\lambda=1$, 1.1 at $\lambda=2$ and 0.56 at $\lambda=4$: that ideal comparison would pick the deeper well with probability 99\%, 87\% and 71\%, and the samples of an actual case, spread over the whole domain, carry less. We call $\lambda\le 1$ the \emph{scored} levels and $\lambda\ge 2$ the \emph{hard} levels; the models are trained on all six, and the hard levels are reported apart.

\subsection{Metrics}
\label{sec:metrics}

\paragraph{Accuracy.} For each case we compute the basins of attraction of the true $f$ on a dense grid, separating neighbouring minima at the ridge between them (watershed basins, Appendix~\ref{app:metrics}). We report the mean and median location error, the share of answers in the right basin (the basin of $\xs$), and the \emph{regret} $(f(\xhat)-f(\xs))/\mathrm{range}(f)$ in percent, which is what a user who evaluates the function at the answer would pay.

\paragraph{Tie-aware measures.} When two valleys are nearly equal, twenty noisy samples cannot tell them apart, and location error cannot distinguish a legitimate choice of the other valley from a wrong answer, nor an answer on a valley floor from one on the ridge between two valleys. For such cases we report whether the answer is \emph{hedging} (more than 5\% of the domain from the floor of every valley; the threshold equals the Huber threshold $\beta$ of the training loss), \emph{in a floor} (within 2\% of one), and, where the geometry is fixed, \emph{on a slope} (closer to a ridge between two basins than to any valley floor, a definition that does not count merely imprecise answers in a single valley). Regret is the summary; on ties it is reported in percent of the barrier between the wells and capped at one barrier, so that a few answers at a steep domain edge cannot dominate it.
\paragraph{Uncertainty.} Every learned answer comes with a per-case spread, which we read as the scale $s$ of a Laplace distribution centred on it; the GP's posterior over the minimiser is scored in the same way. Writing $e=\xhat-\xs$, we report the size of the spread, $\bar s/\overline{|e|}$ (1 is the right size, above 1 too wide); coverage of $|e|\le ks$ for $k=1,2,3$ against $63.2$, $86.5$ and $95.0\%$, and its mean absolute deviation from these targets (\emph{dev}); the probability-integral-transform calibration error (\emph{PIT}, which also applies to full predictive distributions); the expected normalised calibration error over ten spread bins \citep[ENCE,][]{levi2022evaluating}, which catches a spread that is right on average but wrong for confident and doubtful cases; the Laplace negative log-likelihood (NLL) and continuous ranked probability score (CRPS) \citep{gneiting2007strictly}, which reward right size and sharpness together; and two measures of whether the spread knows which answers to doubt: Spearman's $\rho(s,|e|)$, and the excess area under the risk--coverage curve \citep[E-AURC,][]{geifman2019bias}, which keeps answers from most to least confident and measures how far that order is from the best possible one. The point heads of the two Set Transformers were trained on the Huber value of their error rather than on $|e|$ (Section~\ref{sec:baselines}); we read their spread back as the Laplace scale with that expected Huber loss, exactly and with nothing fitted, and use this reading in every comparison. For a two-head model's mode we report the width of the heaviest component; for its mixture mean the spread $\mathbb E|X-\bar x|$ under the mixture; and the full mixture is scored as a distribution. We also report every spread after one cross-fitted scale factor, fitted on alternate cases of each cell and applied to the others, which removes its overall size and leaves only how well it follows the error case by case.

\paragraph{Statistics.} Every estimator answers the same case with the same samples, so all comparisons are paired. Each learned model is the mean over its three seeds. A 95\% percentile-bootstrap interval of a paired difference resamples seeds within each model and cases within each (family, noise level) cell (5{,}000 draws; 2{,}000 for statistics that are not means); on ties the resampling unit is the tie shape, and on the out-of-family sets we also give intervals that resample whole functions, which are the honest ones for a statement about other functions. $^{*}$ marks an interval that excludes zero. Across the thirteen families we also give a Wilcoxon signed-rank test that treats each family as one unit \citep{demsar2006statistical}. The comparisons that carry the paper's claims form one declared family, the $2\times2$ family of Table~\ref{tab:primary}: walk against single pass on in-family regret and error, with one head and with two (point and mode answers), on tie hedging, and on the likelihood of the point spread. For these comparisons we report bootstrap $p$-values with a Holm correction \citep{holm1979simple}, and only claims that survive it are stated as primary. This family was declared after the first seed of each two-head model had been scored on the test cases (Section~\ref{sec:results-selection} gives the order of events); two earlier families, one declared before the comparison of the \nfm{} with the \st{} and one written before the second and third seeds of the two-head models were trained, are reported in Appendix~\ref{app:families-declared}. Appendix~\ref{app:stats} gives the procedures.

\subsection{Evaluation harness and case sets}
\label{sec:harness}

The released harness rebuilds each model from the training source that produced its checkpoint and gives every model inputs built by the input code recorded with it (for all four learned models, the \nfm's); it accepts only the checkpoint at exactly step \trainEpochs{} (never a best-on-validation checkpoint), refuses to run without the reference model, checks the hashes, parameter counts and inputs of every model before scoring, generates every case with the training code's generator, sampler and noise from fixed seeds, applies every exclusion rule before any estimator sees a case, and runs self-checks before scoring (Appendix~\ref{app:harness}). It records library versions, a hash of every checkpoint and a fingerprint of the case set; the standard run reported here was made on a two-core CPU machine and repeated on a Colab CPU with different library versions, with the same case fingerprint and every printed number identical or different by one unit in its last printed digit (Appendix~\ref{app:harness}). The case sets are:
\begin{itemize}\itemsep1pt
\item \emph{in-family}: \nScored{} cases, 13 families $\times$ 4 scored levels $\times$ 100;
\item \emph{hard}: \nHard{} cases, 13 families $\times$ $\lambda\in\{2,4\}$ $\times$ 40;
\item \emph{ties}: \nTies{} cases from 100 double-well shapes whose wells differ by 0, 2 or 10\% of the barrier, at four noise levels;
\item \emph{heavy-tailed}: \nHeavy{} in-family cases with Student-$t$ noise of three degrees of freedom at the same standard deviation as the Gaussian levels $0.25$, $0.5$ and $1$;
\item \emph{unseen}: \nUnseen{} cases from eleven classical test functions never used in development (Section~\ref{sec:results-oof});
\item \emph{monitored}: \nMonitored{} cases from five canonical functions that were printed during development and are therefore not independent evidence.
\end{itemize}
Appendix~\ref{app:casesets} specifies each set to the seed.

\section{The Neural Function Minimizer}
\label{sec:model}

\subsection{Architecture}

\begin{figure}[!t]
\centering
\includegraphics[width=\linewidth]{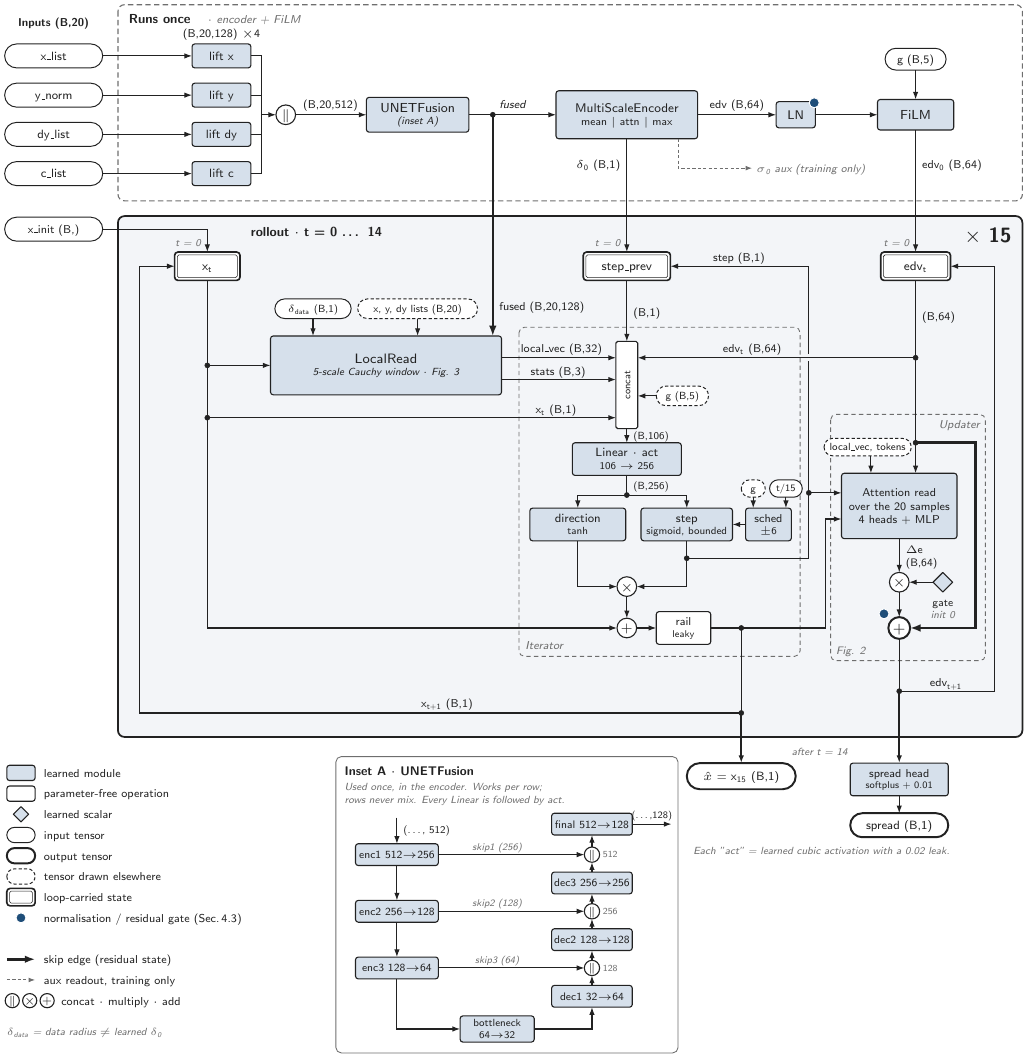}
\caption{The \nfm. Top: the four per-sample channels (with $y$ normalised to $[0,1]$) are lifted, fused position-wise by a U-Net, pooled into the state $\mathbf{e}$, normalised, and conditioned once on the five global descriptors $g$ by FiLM. Bottom, repeated $T=\nIter$ times: the local read returns a Cauchy-windowed summary of the samples around $x_t$ (Figure~\ref{fig:localread}); the iterator maps state, position, previous step, $g$ and the local read to a bounded direction and a bounded step; the updater attends to the twenty samples from the new position and adds a correction to the state through a zero-initialised gate (Figure~\ref{fig:updater}). The answer is $x_T$; the spread head reads the final state. Blue dots mark the normalisation and residual gate of Section~\ref{sec:repair}.}
\label{fig:arch}
\end{figure}

The \nfm{} has four stages (Figure~\ref{fig:arch}).

\paragraph{Scale-free inputs.} The functions of the training families differ in scale by many orders of magnitude, and a model that reads $y$ in its own units can tie its answer to the scale. The \nfm{} therefore receives the shape and the scale of the samples separately (and so do the Set Transformers, from the same code; Section~\ref{sec:baselines}): with $D_y=\max_i y_i-\min_i y_i$, it reads $\tilde y_i=(y_i-\min_j y_j)/D_y\in[0,1]$, computes the spline, its slopes, its coefficients and the descriptors from $\tilde y$, and receives $\log D_y$ as a fifth descriptor. Adding a constant to the twenty observations changes nothing the network sees, and multiplying them by a positive constant changes that one descriptor and nothing else.

\paragraph{Encoder.} Each sample position $i$ carries four channels $(x_i,\tilde y_i,y_i',c_i)$, the location, the normalised value, the spline slope and the $i$-th spline coefficient (Appendix~\ref{app:inputs}); they are embedded channel-wise, the four embeddings are fused by a small U-Net over the sample axis, and a multi-scale pooling produces a 64-dimensional summary vector $\mathbf{e}$ of the whole set, an initial resolution radius $\delta_0$, and an auxiliary noise estimate. The per-sample features $\mathbf{h}_i$ are kept for the local read and the updater. A FiLM layer \citep{perez2018film} conditions $\mathbf{e}$ on the five global descriptors.

\begin{figure}[!t]
\centering
\includegraphics[width=0.96\linewidth]{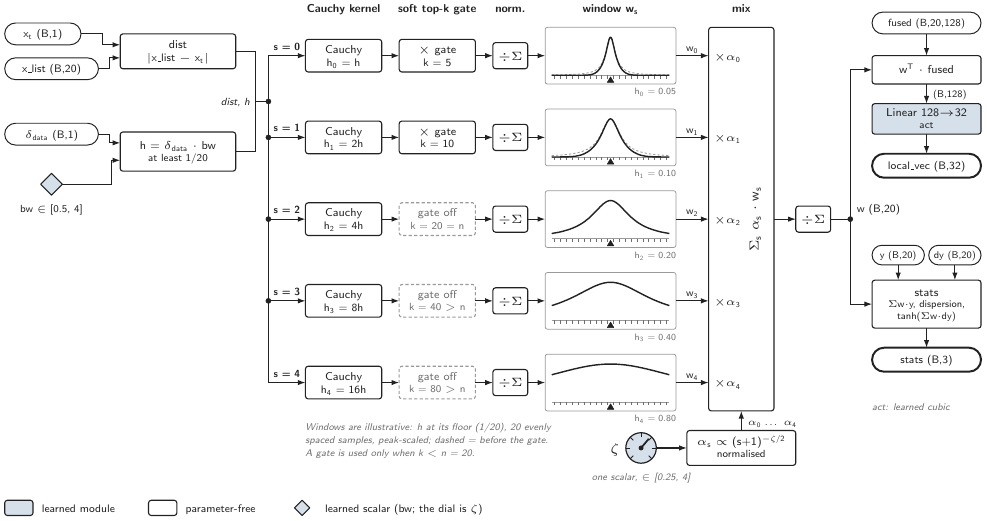}
\caption{The local read. A Cauchy window of width $h=\delta\,\mathrm{bw}$ (floor $1/20$) is evaluated at five dyadic widths, each gated to its $k=5\cdot2^{s}$ nearest samples until the gate covers every sample, normalised, and mixed under a power-law envelope with a single learned exponent $\zeta$. The mixed weights read the fused per-sample features into a 32-dimensional local vector and three statistics.}
\label{fig:localread}
\end{figure}

\paragraph{Local read.} At the current position $x_t$ the model reads the samples through a Cauchy window $1/(1+(d/h)^2)$ of width $h=\delta\,\mathrm{bw}$, where $\delta$ is the resolution radius computed from the data, $\mathrm{bw}\in[0.5,4]$ is a learned multiplier and $h$ has a floor of one sample spacing. The window is evaluated at five dyadic widths $h_s=2^{s}h$, $s=0,\dots,4$, each softly gated to its $k_s=5\cdot2^{s}$ nearest samples (from $s=2$ the gate covers all twenty samples, so the wider reads are global); the five normalised weight vectors are mixed under a power-law envelope $\alpha_s\propto(s+1)^{-\zeta/2}$ with a single learned exponent $\zeta\in[0.25,4]$ (Figure~\ref{fig:localread}). The mixed weights read the fused features $\mathbf{h}_i$ into a 32-dimensional local vector and three statistics (weighted mean of $\tilde y$, dispersion, and the tanh of the weighted slope). In the trained models 65\% of the mixture sits on the narrowest width and the multiplier $\mathrm{bw}$ is at its floor in 72\% of cases, in all three seeds. The width has a floor of one sample spacing, $h=\max(\delta\,\mathrm{bw},1/20)$, so with the multiplier at $0.5$ the window sits at its floor whenever $\delta<0.1$, which is most cases: in practice the narrowest read covers about one sample spacing.

\paragraph{Iterator.} A single hidden layer of 256 units reads $[\mathbf{e},x_t,\eta_{t-1},\mathbf{g},\text{local}]$, where $\eta_{t-1}$ is the previous step size ($\eta_0=\delta_0$, the encoder's initial radius), and emits a direction in $(-1,1)$ and a bounded step size $\eta_t$; a schedule conditioned on $\mathbf{g}$ and $t/T$ shifts the step's pre-activation so that the total travel over $T=\nIter$ steps is at most twice the domain. The position is updated, $x_{t+1}=x_t+\eta_t\cdot\text{direction}$.

\begin{figure}[!t]
\centering
\includegraphics[width=0.92\linewidth]{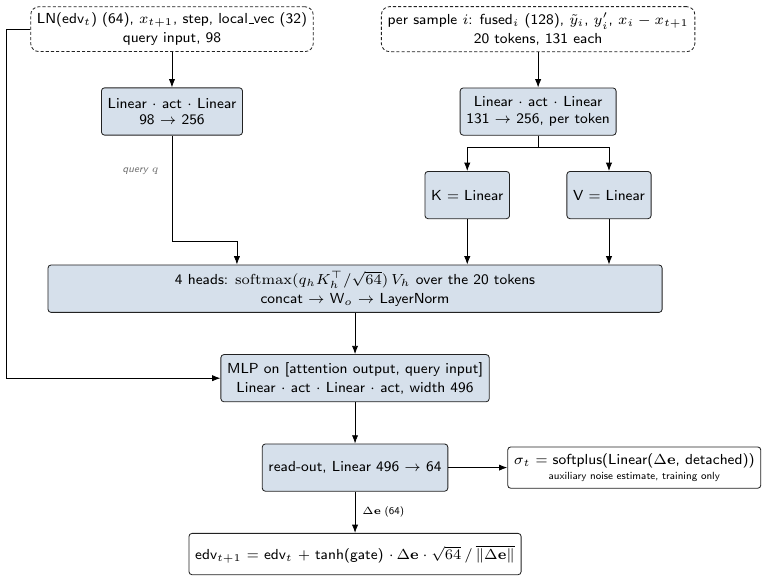}
\caption{The updater. The query is built from the normalised state, the new position, the step and the local vector; each of the twenty samples becomes a token carrying its fused encoder features, its normalised value, its slope and its offset from the new position. Four attention heads read the tokens; a two-layer network combines the result with the query and emits a 64-dimensional correction, which is added to the state through the gate after normalisation by its running norm. A detached auxiliary head estimates the noise during training only.}
\label{fig:updater}
\end{figure}

\paragraph{Updater.} The updater revises the state from the new position by reading the samples (Figure~\ref{fig:updater}). Its query is built from $[\mathrm{LN}(\mathbf e_t),x_{t+1},\eta_t,\text{local vector}]$; each sample is a token $[\mathbf h_i,\tilde y_i,y_i',x_i-x_{t+1}]$, so every token carries its position relative to where the walk now stands. Both are mapped to 256 dimensions; four attention heads \citep{vaswani2017attention} read the twenty tokens, followed by an output projection, layer normalisation \citep{ba2016layer}, a two-layer network of width 496 on the attention output and the query, and a read-out to a 64-dimensional correction $\Delta\mathbf e$. The state is updated residually, $\mathbf{e}\leftarrow\mathbf{e}+\tanh(\gamma)\,\Delta\mathbf{e}\,\sqrt{64}/\overline{\lVert\Delta\mathbf e\rVert}$, with a scalar gate $\gamma$ initialised at zero \citep{bachlechner2021rezero} and $\overline{\lVert\Delta\mathbf e\rVert}$ a running mean of the correction's norm, so that the gate alone sets the size of the update. The updater holds 63.6\% of the parameters (\paramsUpdater{} of \paramsNFMexact).

\paragraph{Activation.} Every hidden unit applies a learned combination of the first three truncated powers of its pre-activation, $\psi(x)=a_1 r+a_2 r^{2}+a_3 r^{3}$ with $r=\mathrm{clip}\big(\mathrm{leaky}_{0.02}(x),-0.2,10\big)$ and $a_1,a_2,a_3>0$ learned per layer. The powers $(x)_+$, $(x)_+^2$ and $(x)_+^3$ are the local basis of linear, quadratic and cubic splines, and networks built from such rectified power units approximate smooth functions more efficiently than ReLU networks and represent low-degree polynomials exactly \citep{li2020better}; learned activations of this kind have precedents in Pad\'e activation units \citep{molina2020pade}, rational networks \citep{boulle2020rational} and the learned B-spline activations of Kolmogorov--Arnold networks \citep{liu2025kan}. The clamp bounds the growth of the cubic, and the small slope below zero keeps a unit's gradient alive for pre-activations in $(-10,10)$; both were added after we observed saturated and dead units (Section~\ref{sec:repair}). We did not ablate the activation and make no claim that it drives the results.

\paragraph{Outputs.} The answer is $x_T$. A linear head on the final $\mathbf{e}$, followed by a softplus plus 0.01 (so that $s\ge0.01$), gives the Laplace scale $s$. The model has \paramsNFMexact{} parameters.

\subsection{The second head}
\label{sec:twohead}

The two-head walk, \nfmmh, is the \nfm{} with one more linear layer on its final state $\mathbf e$: a map from 64 to 15 numbers that give, for $k=1,\dots,5$, a weight $w_k$ (softmax), a location $\mu_k\in(0,1)$ (sigmoid) and a Laplace scale $b_k$ (softplus plus 0.01). The point head, the spread head and every other part of the \nfm{} are unchanged, and so are its losses; the negative log-likelihood of the label under the mixture is added with weight 1.0, down-weighted to $0.3$ on unresolvable cases like the position loss, and its gradient reaches the whole network. The second head adds 975 parameters (\paramsNFMMHexact{} in all). The model answers in three ways: the walk's endpoint $x_T$ (\emph{point}, with the spread head's $s$), the location $\mu_{k^*}$ of the heaviest component (\emph{mode}, with its scale $b_{k^*}$), and the mixture mean $\bar x=\sum_k w_k\mu_k$ (with the spread $\mathbb E|X-\bar x|$ under the mixture). Its single-pass counterpart, \sttwo, carries the same head with the same initialisation and loss (Section~\ref{sec:baselines}).

\subsection{Training}

Cases are generated on the fly from the thirteen families, uniformly over the six noise levels. Half of the rollouts start at the spline argmin and half at a uniformly random point, so the model must be able to relocate. The loss is a Huber loss (threshold $\beta=0.05$) on the final position plus a trajectory term that weights the position error at step $t$ by $t$, a Laplace calibration term $|e|/s+\log s$ in which the error is detached so that $s$ cannot be widened to cheapen the position loss, an auxiliary noise-supervision term, and a penalty on positions outside $[0,1]$; the \nfmmh{} adds the mixture term. Cases with no sample in the true basin are down-weighted to $0.3$. Training runs \trainEpochs{} steps with batch size 48 and Adam at learning rate $2\times10^{-4}$ on a cosine schedule over $3\times10^6$ steps, which is still above $1.9\times10^{-4}$ at the end, with gradient clipping at 1.0 and a guard that skips a batch whose gradient norm exceeds 100 times the running median of accepted norms. Runs were trained in one to three sessions, each later session resumed from a saved checkpoint, with seeds drawn at random and recorded; no run needed a rollback. No training run was discarded: the three runs of each model are the only runs of that model we started. Every model was trained on Google Colab on one NVIDIA L4 GPU (high-RAM runtime). Each training run of a walk model took 33 to 34 hours of wall-clock time, and each Set Transformer run 3.2 to 3.4 hours; these are the times from the first to the last evaluation in each run's log, summed over its sessions. Appendix~\ref{app:hyper} lists the hyperparameters.

\subsection{Keeping the update module alive}
\label{sec:repair}

The updater is the largest module of the model and the one most easily lost. During development, an earlier design trained to the same loss and to a plausible error while not using its updater at all: the pre-activations of the clamped cubic activation inside the module had grown to magnitudes of 13 to 43 against a clamp at 10, every unit was dead or pinned, and the module emitted the same vector for every input. Replacing its output by that constant changed no prediction (maximum change in any answer $<10^{-6}$); the state vector $\mathbf{e}$ had zero variance across cases after the first step; and the uncertainty head, which reads $\mathbf{e}$, gave the same number for every case. The model had become a single-pass network with a large inert module attached, and nothing in the logged losses showed it.

The final design follows from this. The state is normalised where it is produced and at the updater's input, so that every consumer reads values of order one; the update is residual with a zero-initialised gate and a normalised correction, so that at initialisation the model is a working single-pass network and the updater can only add to the state; the activation keeps a small slope (0.02) below zero so that no unit loses its gradient; the calibration term is trained on $|e|$ rather than on the Huber value of $e$, so that the spread reports the model's own typical error; and the updater reads the samples directly instead of a single compressed vector. The training pipeline runs a liveness monitor at every evaluation (state variance across cases, live-unit fractions per layer, a frozen-correction ablation, and the calibration of the spread head). In the three \nfm{} runs, 1.0 to 2.7\% of the updater's units never fire, the state's standard deviation across cases grows from 0.56--0.58 at the first step to 0.77--0.90 at the last, and freezing the updater's correction at its case average raises the error by 6.4 to 11.4 points of the domain on the monitor's 480 fixed in-family cases (Appendix~\ref{app:ablations}); in the three \nfmmh{} runs, 0.3 to 0.6\% never fire, the state's standard deviation grows from 0.32--0.38 to 0.47--0.54, and freezing the correction raises the error by 6.3 to 8.9 points.

\section{Baselines}
\label{sec:baselines}

\paragraph{Set Transformer (\st).} The reference architecture of \citet{lee2019set}, with its attention blocks (MAB, SAB, ISAB, PMA) taken verbatim from the authors' implementation. Each of the 20 observations is a four-dimensional token $(x_i, \tilde y_i, y_i', c_i)$; two induced set-attention blocks (4 heads, 32 inducing points) encode the set, pooling by multi-head attention (one seed) gives one vector, the five global descriptors $\mathbf g$ and the resolution radius $\delta$ are concatenated to it, two self-attention blocks decode it into a vector $H$, and a linear head on $H$ gives $\xhat$ through a sigmoid and $s$ through a softplus. Layer normalisation is off, as in the reference default. The hidden width is 228 rather than the reference 128, chosen to match the \nfm's parameter budget: \paramsSTexact{} parameters against the \nfm's \paramsNFMexact{} (0.2\% more), both counted from the saved checkpoints. It is trained on the identical batch generator, family mixture, noise levels (the six levels with equal probability, as recorded in every checkpoint) and samplers, with the same optimiser, schedule, gradient clipping and gradient guard, for the same \trainEpochs{} steps, with the final-position Huber loss; the three loss terms that only make sense for a rollout (trajectory, out-of-bounds, auxiliary noise) are dropped. Its inputs are the \nfm's: the same scale-free tokens, the same five descriptors and the same resolution radius, built by the \nfm's own input code, so that the two models differ in architecture and loss but not in what they read. The one input it does not use is the \nfm's starting point, the spline argmin, which a single-pass model has no use for; the walk's answer depends little on it (Appendix~\ref{sec:results-path}). Its calibration term fits the scale to the Huber value of the error, so its raw spread estimates $\mathbb E[\mathrm{Huber}(e)]$ rather than $\mathbb E|e|$; the harness reads it back to the Laplace scale exactly (Section~\ref{sec:metrics}).

\paragraph{Two-head Set Transformer (\sttwo).} The \st{} with the mixture head of the \nfmmh{} added on its decoder output $H$, which is also the input of its point head: one linear layer from 228 to 15 numbers, with the same parametrisation, the same initialisation, the same mixture loss with weight 1.0 and the same down-weighting of unresolvable cases, and its gradient reaching the whole network. Its point head and its losses are the \st's. It answers in the same three ways. It has \paramsSTTWOexact{} parameters. The asymmetries that remain between the two two-head models are those of the base models: the walk's point loss has terms that a single pass cannot have, the two spread heads are trained to different targets ($|e|$ and the Huber value of $e$), and the mixture head reads a 64-dimensional state in the walk and a 228-dimensional vector in the single pass.

\paragraph{Looped Set Transformer (\stl).} A control for the amount of computation, added last (Section~\ref{sec:results-selection}). It is the \st{} with its decoder's second block replaced by a loop of 32 steps with shared weights: each step re-reads the encoded samples $Z$ by attention and revises one state vector, $H\leftarrow\mathrm{SAB}(\mathrm{MAB}(H,Z))$, with layer normalisation in both looped blocks. There is no position and no offset; the loop repeats a read of all the samples, as the walk's updater does, without moving. The number of steps was set so that its operations per training step equal the \nfm's: a forward and backward pass on one training batch of 48 costs 24.7 GFLOP, against 24.6 for the \nfm{} and 6.4 for the \st. Its width is 212, giving \paramsSTLexact{} parameters. Data, inputs, loss, optimiser, schedule, gradient guard and the \trainEpochs{} steps are the \st's. It was trained in one configuration, three seeds, each in 15.9 to 16.8 hours.

We trained one configuration of each Set Transformer, sized to match the \nfm, with no hyperparameter sweep, whereas the \nfm's design was chosen among several variants during development (Section~\ref{sec:results-selection}); the comparison favours the walk in this respect, and a tuned Set Transformer could do better. The two-head models and the \stl{} were each trained in one configuration, decided before their first run. Three seeds were trained of each learned model.

\paragraph{Classical zero-query estimators.} The GP posterior argmin, with a Mat\'ern-5/2 kernel plus white noise, hyperparameters by maximum marginal likelihood and the argmin taken between the first and last sample \citep{rasmussen2006gaussian}; from the same posterior, 256 draws of the latent function give the posterior distribution of the argmin \citep{hennig2012entropy}, whose median (the best guess under absolute error) and mode (which commits to one valley) are two further estimators and whose spread is scored like the learned models' (with the same floor of $0.01$); a local quadratic regression with its bandwidth chosen by leave-one-out cross-validation, the standard plug-in estimator for this problem \citep{muller1989adaptive} and a moving-weight local polynomial fit of the kind analysed by \citet{levin1998approximation}; the argmin of a smoothing spline fitted to the raw samples by the routine that produces the learned models' inputs \citep{wahba1990spline} (the learned models' spline is fitted to the normalised samples, and because the smoothing has an absolute floor the two splines can differ, so the baseline is not exactly the walk's starting point); differential evolution \citep{storn1997differential} and basin-hopping \citep{wales1997global} run on that spline; and the lowest observation.

\paragraph{Ensemble.} The median of the \nfm, \st{} and GP posterior-argmin answers, which costs no extra model call.

\section{Results}
\label{sec:results}

\subsection{Order of events}
\label{sec:results-selection}

To separate confirmatory from exploratory evidence, we state the order in which things were done. The \nfm's design was developed over several variants, monitored on in-family evaluation cases and on the five monitored functions of Section~\ref{sec:results-oof}; the choice between its last two designs was made after seeing their test results, and a run on validation cases with a separate seed block, which enter no reported result, agreed with it. A first comparison of the \nfm{} with two Set Transformers, one with a point head and one with only a mixture head, and the classical estimators was made with a list of primary comparisons declared before it was run. Those Set Transformers read $y$ in its own units; because that mixed a difference in inputs with one in architecture, both were retrained on the \nfm's inputs and the same comparisons were scored again on the same cases (Appendix~\ref{app:families-declared}). The \st{} of this paper is the retrained point model. Its mixture-only model differed from the \nfm{} in both its architecture and the form of its output. To separate the two, we then trained the \nfmmh, the \nfm{} with a second, mixture head. Its first seed was scored on the test cases at five points of its training (77k, 125k, 240k, 306k and 420k steps). Before its second and third seeds were trained, we wrote down an analysis plan of five comparisons. Because the \nfmmh{} kept its point head while the mixture-only model had none, we then trained the \sttwo, which carries the same two heads, and the $2\times2$ family of Table~\ref{tab:primary} was declared after the first seed of both two-head models had been scored. The second and third seeds of both were scored once, at step \trainEpochs, in the run reported here. Two comparisons of the analysis plan were against the mixture-only Set Transformer, which this paper replaces by the \sttwo; Appendix~\ref{app:families-declared} reports both earlier families in full, including every comparison with the mixture-only model and both runs of the first family. The split of regret by basin, the median regret, the moving-viewpoint probe and the table of hedging by noise level (Table~\ref{tab:hedgenoise}) were added after all seeds had been scored, and are exploratory. The \stl{} was added last, to test whether the walk's lower regret comes from repeated computation. Its four comparisons, and how each outcome would be read, were written down before any looped model was trained. A quick run of the harness, on 2{,}080 cases that partly coincide with the test cases (160 share a function and sample locations with a test case), was seen before the standard run. The wording of the reading was narrowed after the standard results were seen, and the analyses of the loop's steps and every comparison outside the four are exploratory; the plan and its two dated addenda are in the repository. Finally, every declared comparison of Table~\ref{tab:primary} and Table~\ref{tab:looped} was run once more, with the same trained models, on a block of cases that no one had seen, under a plan written before that run (SHA-256 \texttt{c4520eb9b6359b53\-90662724d55a0877\-11c0cafa29feb88d\-14030f4e998ccb22}); an addendum refining two of its reading rules was written after the run had finished and before its output was analysed (SHA-256 \texttt{c031383cffc27bfc\-b410266f9848807c\-ab3c3f7e734bdbd6\-0eb0c4aaa4cbe850}), and it changes no comparison and no conclusion (Section~\ref{sec:results-fresh}). The eleven unseen functions entered no development decision. The tie effect was first noticed on Griewank during development, so the Griewank result is exploratory and the constructed tie set is the test of it.

\subsection{Accuracy}
\label{sec:results-accuracy}

\begin{table}[t]
\centering
\footnotesize
\setlength{\tabcolsep}{2.4pt}
\begin{tabular}{lrrrrll}
\toprule
& \multicolumn{2}{c}{Error} & Right & & \multicolumn{2}{c}{Method $-$ \nfm} \\
\cmidrule(lr){2-3}\cmidrule(lr){6-7}
Method & Mean & Median & basin & Regret & Error & Regret \\
\midrule
\nfm{} (walk)                      & 6.72\,{\scriptsize$\pm$0.07} & 2.31 & 82.4\% & 12.80 & --- & --- \\
\nfmmh{} (walk), point             & \textbf{6.27}\,{\scriptsize$\pm$0.11} & 1.71 & 82.9\% & 10.09 & $-0.44$ [$-0.64$, $-0.27$]$^{*}$ & $-2.71$ [$-3.35$, $-2.07$]$^{*}$ \\
\quad mode                          & 6.35\,{\scriptsize$\pm$0.13} & \textbf{1.06} & 82.6\% & 6.53 & $-0.36$ [$-0.60$, $-0.14$]$^{*}$ & $-6.27$ [$-6.88$, $-5.65$]$^{*}$ \\
\quad mixture mean                  & 6.86\,{\scriptsize$\pm$0.11} & 2.81 & 82.4\% & 17.65 & $+0.14$ [$-0.05$, $+0.32$] & $+4.85$ [$+3.87$, $+6.01$]$^{*}$ \\
\st{} (single pass)                & 6.74\,{\scriptsize$\pm$0.02} & 2.84 & 82.8\% & 15.70 & $+0.02$ [$-0.14$, $+0.18$] & $+2.90$ [$+2.13$, $+3.78$]$^{*}$ \\
\sttwo{} (single pass), point & 6.28\,{\scriptsize$\pm$0.01} & 2.09 & \textbf{83.4\%} & 13.21 & $-0.43$ [$-0.60$, $-0.27$]$^{*}$ & $+0.41$ [$-0.16$, $+0.99$] \\
\quad mode                          & 6.37\,{\scriptsize$\pm$0.10} & 1.44 & 82.9\% & 7.98 & $-0.35$ [$-0.57$, $-0.13$]$^{*}$ & $-4.81$ [$-5.34$, $-4.28$]$^{*}$ \\
\quad mixture mean                  & 6.86\,{\scriptsize$\pm$0.08} & 3.02 & 82.3\% & 18.35 & $+0.15$ [$-0.04$, $+0.33$] & $+5.56$ [$+4.74$, $+6.33$]$^{*}$ \\
\midrule
GP posterior argmin               & 8.02 & 1.40 & 78.8\% & \textbf{5.53} & $+1.30$ [$+1.00$, $+1.60$]$^{*}$ & $-7.27$ [$-7.78$, $-6.75$]$^{*}$ \\
GP argmin posterior, median       & 7.79 & 1.66 & 78.6\% & 7.54 & $+1.08$ [$+0.79$, $+1.37$]$^{*}$ & $-5.26$ [$-5.81$, $-4.69$]$^{*}$ \\
GP argmin posterior, mode         & 9.04 & 1.69 & 77.8\% & 6.73 & $+2.32$ [$+1.99$, $+2.66$]$^{*}$ & $-6.07$ [$-6.61$, $-5.50$]$^{*}$ \\
Local quadratic (LOOCV)           & 11.10 & 3.57 & 72.3\% & 13.00 & $+4.39$ [$+4.00$, $+4.78$]$^{*}$ & $+0.20$ [$-0.42$, $+0.83$] \\
Lowest observation                & 10.06 & 2.99 & 75.2\% & 7.78 & $+3.35$ [$+3.04$, $+3.66$]$^{*}$ & $-5.02$ [$-5.53$, $-4.50$]$^{*}$ \\
Diff.\ evolution on spline       & 13.21 & 5.38 & 68.8\% & 16.18 & $+6.49$ [$+6.08$, $+6.90$]$^{*}$ & $+3.38$ [$+2.73$, $+4.05$]$^{*}$ \\
Basin-hopping on spline           & 14.05 & 5.75 & 67.2\% & 16.02 & $+7.33$ [$+6.91$, $+7.77$]$^{*}$ & $+3.22$ [$+2.58$, $+3.88$]$^{*}$ \\
Spline argmin                     & 14.21 & 5.75 & 66.9\% & 16.10 & $+7.49$ [$+7.06$, $+7.94$]$^{*}$ & $+3.30$ [$+2.63$, $+3.98$]$^{*}$ \\
\bottomrule
\end{tabular}
\caption{Location error (percent of the domain) and regret (percent of the function's range) on the \nScored{} in-family cases (13 families $\times$ $\lambda\in\{0,0.25,0.5,1\}$ $\times$ 100). Every method answers from the same 20 samples of the same case. The \nfm{} and the \st{} have one output head; the \nfmmh{} and the \sttwo{} add a second, mixture head. A two-head model gives three answers from one network: its point head (\emph{point}), the location of its heaviest mixture component (\emph{mode}) and the weighted mean of its components (\emph{mixture mean}). Learned models are means over three training seeds ($\pm$ the standard deviation of the seed means). The last two columns are paired differences, the row's method minus the \nfm, with 95\% intervals from a bootstrap that resamples seeds within each model and cases within each (family, level) cell; negative values favour the row's method, and $^{*}$ marks an interval that excludes zero. Bold marks the best value in each column. Every surrogate argmin is taken between the first and last sample. The median of the \nfm, \st{} and GP answers (not shown) reaches 6.16 (Appendix~\ref{sec:results-ensemble}).}
\label{tab:benchmark}
\end{table}

\begin{table}[t]
\centering\small
\setlength{\tabcolsep}{4pt}
\begin{tabular}{llrrr}
\toprule
Measure & Comparison (walk $-$ single pass) & Difference [95\% CI] & $p$ & Holm $p$ \\
\midrule
In-family regret & \nfmmh{} $-$ \sttwo, point & $-3.12$ [$-3.78$, $-2.49$] & $<$0.0002 & \textbf{$\le$0.0016} \\
In-family error & \nfmmh{} $-$ \sttwo, point & $-0.01$ [$-0.19$, $+0.17$] & 0.91 & 1 \\
In-family regret & \nfmmh{} $-$ \sttwo, mode & $-1.45$ [$-1.93$, $-0.98$] & $<$0.0002 & \textbf{$\le$0.0016} \\
In-family error & \nfmmh{} $-$ \sttwo, mode & $-0.01$ [$-0.27$, $+0.25$] & 0.93 & 1 \\
Tie hedging (points) & \nfmmh{} $-$ \sttwo, point & $+2.4$ [$-4.4$, $+9.4$] & 0.52 & 1 \\
NLL & \nfmmh{} $-$ \sttwo, point (read-back) & $-0.12$ [$-0.19$, $-0.06$] & $<$0.0005 & \textbf{$\le$0.0025} \\
In-family regret & \nfm{} $-$ \st & $-2.90$ [$-3.76$, $-2.14$] & $<$0.0002 & \textbf{$\le$0.0016} \\
In-family error & \nfm{} $-$ \st & $-0.02$ [$-0.19$, $+0.13$] & 0.77 & 1 \\
\bottomrule
\end{tabular}
\caption{The $2\times2$ family of declared comparisons, walk minus single pass (error in points of the domain, regret in points of the range, hedging in percentage points of the tie cases, NLL in nats); negative values favour the walk (for tie hedging, a negative value means the walk hedges less). Bold: significant after Holm's correction over the eight comparisons. Holm's adjusted values are monotone in the ordered $p$, so tied values occur. A bootstrap $p$ cannot be smaller than one over the number of draws: 0.0002 for the means (5{,}000 draws) and 0.0005 for the NLL (2{,}000 draws, as for every calibration measure); where the observed $p$ is at that floor, the true $p$ and the Holm value are bounded above, shown by $<$ and $\le$. The family was declared after the first seed of each two-head model had been scored (Section~\ref{sec:results-selection}). The intervals come from this family's own bootstrap draws and can differ from those of the same comparison elsewhere in the paper in the second decimal.}
\label{tab:primary}
\end{table}

Table~\ref{tab:benchmark} gives location error and regret on the in-family cases. Location error is bimodal: an answer in the right basin is usually close and one in the wrong basin far, so the mean (6.72 for the \nfm) lies well above the median (2.31), and the mean mixes how often an estimator chooses the right basin with how precisely it locates the minimum inside it. We therefore read the right-basin rate, the median and regret alongside the mean.

\paragraph{The walk and the single pass are tied on the mean.} With one head, the \nfm{} reaches 6.72 (seeds 6.67, 6.68 and 6.80) and the \st{} 6.74 ($-0.02$ [$-0.19$, $+0.13$], walk minus single pass); with two heads, the point answers reach 6.27 and 6.28 ($-0.01$ [$-0.19$, $+0.17$]) and the modes 6.35 and 6.37; none of these differences is significant (Table~\ref{tab:primary}). The walk and the single pass choose the right basin equally often (walk minus single pass $-0.3$ [$-1.0$, $+0.3$] points with one head and $-0.5$ [$-1.4$, $+0.3$] with two; the four models lie between 82.4 and 83.4\%). The walk is ahead on the median error: 2.31 against 2.84 with one head, 1.71 against 2.09 for the point answers with two ($-0.37$ [$-0.50$, $-0.23$]) and 1.06 against 1.44 for the modes ($-0.38$ [$-0.45$, $-0.30$]). With equal means and equal right-basin rates, a lower median means that the walk's answers in the right basin lie closer to the minimum, and its answers in the wrong basin, on average, farther.

\paragraph{The mixture loss lowers the error, in both designs.} Adding the mixture head and its loss lowers the mean error of the point answer by about the same amount in both designs: 0.44 [0.27, 0.64] points for the walk (6.72 to 6.27) and 0.45 [0.33, 0.58] for the single pass (6.74 to 6.28); the difference between the two effects is $+0.01$ [$-0.21$, $+0.23$]. It lowers regret by 2.71 and 2.49 points. This gain belongs to the loss, not to the walk; across the thirteen families as units the \nfmmh's point answer is better than the \nfm{} in all thirteen (Wilcoxon $p=0.0002$). The two-head models' mixture means are less accurate than their point answers (6.86 in both) and have the highest regret of any learned answer (17.65 and 18.35), for a reason Section~\ref{sec:results-tie} makes plain.

\paragraph{The classical estimators.} Every learned answer, the mixture means included, is more accurate on the mean than every classical estimator, the GP included: the GP posterior argmin reaches 8.02 (\nfm{} better by 1.30 [1.00, 1.60] points), the median of the GP's argmin posterior 7.79, the local quadratic regression 11.10. The advantage is on the mean and on the right-basin rate (78.8\% for the GP): the GP has a lower median error than every learned point answer (1.40), and the lowest regret of all (5.53), below the best learned answer, the \nfmmh's mode (6.53); across families as units the \nfm's lead over the GP is not significant ($p=0.24$, ahead in 9 of 13 families). Within the training families the learned models and the GP are therefore better described as competitive, each ahead on different measures, than as ranked; the GP is also a single standard configuration, not one tuned to these families.

\paragraph{Hard levels.} At $\lambda=2$ and $4$, where the data do not determine the answer, the walk and the single pass are tied on error (\nfm{} 13.67, \st{} 13.41, \nfmmh{} 13.03, \sttwo{} 13.33; walk minus single pass $+0.26$ [$-0.26$, $+0.76$] with one head and $-0.31$ [$-0.72$, $+0.12$] with two), the mixture loss lowers the walk's error here too (\nfmmh{} $-$ \nfm{} $-0.65$ [$-1.14$, $-0.16$]), and all four are ahead of the GP (17.22). The walk's regret is lower here too, by 2.89 [1.72, 4.08] with one head and 3.22 [1.79, 4.62] with two (Table~\ref{tab:walk}); the modes (19.01 and 19.36) and the GP (17.77) have lower regret still.

\paragraph{Robustness slices.} Under heavy-tailed noise the learned models stay close (\nfm{} 7.05, \nfmmh{} 6.63, \st{} 6.85, \sttwo{} 6.37; GP 8.26), and the walk's regret remains lower with two heads ($-2.40$ [$-3.58$, $-1.14$] for the point answers, $-1.02$ [$-1.84$, $-0.19$] for the modes). The errors under Student-$t$ noise are lower than or equal to those under Gaussian noise of the same standard deviation for every estimator (ratios 0.87 to 1.00); the two sets are drawn with different seeds and so contain different functions, and the comparison is not paired. By sampling design (uniform, Gaussian, Beta) the two-head models are ahead of the one-head models in every design, and all four are ahead of every classical estimator.

\subsection{What the walk changes}
\label{sec:results-walk}

\begin{figure}[t]
\centering
\includegraphics[width=\linewidth]{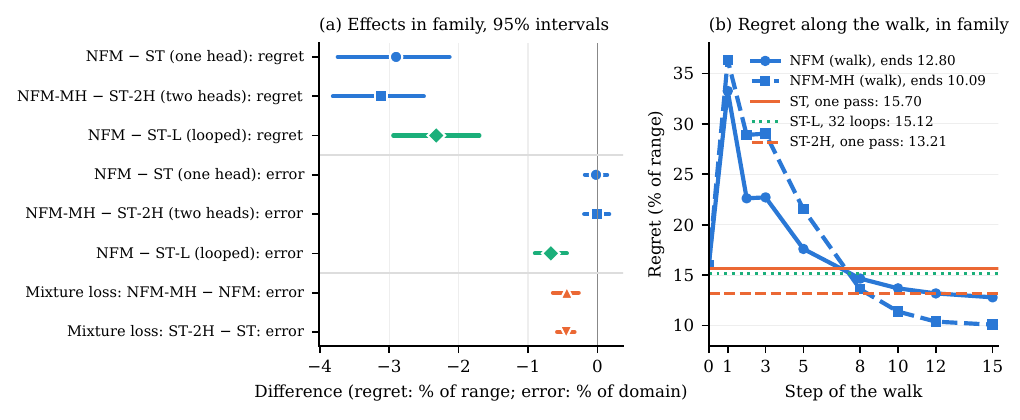}
\caption{The main results in family. (a) Paired differences with 95\% intervals (regret in points of the range, error in points of the domain; negative favours the first model): the walk's regret against the single pass with one head and with two (Table~\ref{tab:walk}) and against the looped Set Transformer (Table~\ref{tab:looped}), the same for error, and the mixture loss's effect on error in each design (Section~\ref{sec:results-accuracy}). (b) Regret of the walk's position after each step (Table~\ref{tab:path}), against the one answer of each single-pass model; three seeds each.}
\label{fig:summary}
\end{figure}

\begin{table}[t]
\centering\footnotesize
\setlength{\tabcolsep}{3pt}
\begin{tabular}{llrrr}
\toprule
Case set & Measure & One head: \nfm{} $-$ \st & Two heads: \nfmmh{} $-$ \sttwo & Change \\
\midrule
In family & Error  & $-0.02$ [$-0.18$, $+0.14$] & $-0.01$ [$-0.19$, $+0.17$] & $+0.01$ [$-0.21$, $+0.23$] \\
          & Regret & $-2.90$ [$-3.74$, $-2.13$]$^{*}$ & $-3.12$ [$-3.81$, $-2.50$]$^{*}$ & $-0.23$ [$-1.26$, $+0.83$] \\
Hard levels & Error  & $+0.26$ [$-0.26$, $+0.76$] & $-0.31$ [$-0.72$, $+0.12$] & $-0.57$ [$-1.15$, $+0.02$] \\
          & Regret & $-2.89$ [$-4.08$, $-1.72$]$^{*}$ & $-3.22$ [$-4.62$, $-1.79$]$^{*}$ & $-0.33$ [$-2.09$, $+1.44$] \\
Unseen functions & Error  & $-0.29$ [$-0.94$, $+0.34$] & $+0.18$ [$-0.36$, $+0.69$] & $+0.47$ [$-0.34$, $+1.29$] \\
          & Regret & $-2.48$ [$-4.23$, $-0.82$]$^{*}$ & $-1.27$ [$-2.94$, $+0.39$] & $+1.21$ [$-1.11$, $+3.62$] \\
\bottomrule
\end{tabular}
\caption{The effect of the walk, point answers: walk minus single pass with one head and with two (error in points of the domain, regret in points of the range), and the change between the two (two heads minus one head). Negative values favour the walk. Each bootstrap replicate resamples the cases once, shared by the four models, within each (family, level) cell, and the seeds within each model (on the unseen functions, too: cases, not functions; Section~\ref{sec:results-oof} gives the intervals with whole functions resampled); $^{*}$ marks an interval that excludes zero. The change is also the difference between the mixture loss's effect on the walk and on the single pass.}
\label{tab:walk}
\end{table}

Figure~\ref{fig:summary} summarises the main effects. Table~\ref{tab:walk} gives the walk's effect on the point answer with one head and with two. In family it is the same in both: no effect on the mean error, and a regret lower by 2.90 and 3.12 points, the change between the two being $-0.23$ [$-1.26$, $+0.83$]. On the hard levels it is the same again, and on the unseen functions the walk's lower regret, significant with one head, is not significant with two. In family the effect is about the same with either head: the change interval, $-1.26$ to $+0.83$, excludes differences larger than about 40\% of the effect. On the hard levels and the unseen functions the change intervals are as wide as the effect itself.

Is the lower regret a property of the wrong answers only? Split by basin, it is not. On the in-family cases the walk's answers in the wrong basin cost less (regret 30.5 against 35.4\% of the range with one head, $-4.9$ [$-7.1$, $-2.7$]; 26.6 against 32.7 with two, $-6.1$ [$-8.1$, $-4.0$]), and so do its answers in the right basin (9.0 against 11.6, $-2.6$ [$-3.4$, $-1.9$]; 6.7 against 9.3, $-2.7$ [$-3.3$, $-2.0$]); the share of answers in the wrong basin is the same (17.6 and 17.2\%; 17.1 and 16.6\%). The median regret is lower as well (2.23 against 3.09 with one head, 1.35 against 1.97 with two). The walk's answers lie closer to the floor of whichever valley they choose. Because most answers are in the right basin, about three quarters of the regret advantage comes from them (about 2.1 of 2.90 points with one head and 2.2 of 3.12 with two, the right-basin difference weighted by the share of answers there). (Intervals: cases resampled within cells and seeds within models, 4{,}000 draws; the conditional means are not paired, because the two models' answers fall in the wrong basin on different cases.)

Is the effect carried by one seed or by a few families? Every seed of the \nfm{} has lower regret than every seed of the \st{} (nine seed pairs; differences $-2.23$ to $-4.06$), and the same holds for every seed of the \nfmmh{} against every seed of the \sttwo{} ($-2.48$ to $-3.97$) and for their modes ($-0.93$ to $-2.00$); Appendix~\ref{app:seeds} gives every seed. Across the 13 families the walk's regret is lower in 10, with one head and with two (exact two-sided Wilcoxon test over families, $p=0.005$ and $0.003$). The difference is near zero on the bowls, the double wells and the splines ($+0.04$ to $+0.38$ with either head) and largest on the plateau valleys and the decaying ripples ($-7.7$ to $-13.5$). These checks were computed after all seeds had been scored.

Where along the walk does the lower regret arise? The walk's position can be scored after every step (Table~\ref{tab:path}). Both walk models start from the argmin of the spline in their inputs, whose regret is 16.0\% of the range (the spline-argmin baseline, fitted to the raw samples, has 16.10). After one step the regret jumps to 33.3 (\nfm) and 36.3 (\nfmmh), and the share of tie cases on which the position hedges between valleys to 92 and 96\%: the first steps move the position into the middle of the domain. The walk then descends. The \nfm's regret falls below the \st's single answer (15.70) by step 8 and reaches 12.80 at step 15; the \nfmmh's falls below the \sttwo's (13.21) by step 10 and reaches 10.09. Its hedging on ties falls over the same steps, from 96\% at step 1 to 22.6\% at step 15, about where the single pass ends (20.2\%). The mean error is within 5\% of its final value by step 8 for the \nfm{} and step 11 for the \nfmmh{} (the harness computes this from every step; the table shows selected steps). The lower regret is thus earned in the second half of the walk: the steps after the error has nearly settled move the answers that are already in a valley closer to its floor.

\begin{table}[t]
\centering\small
\setlength{\tabcolsep}{3.5pt}
\begin{tabular}{llrrrrrrrrr}
\toprule
& & \multicolumn{9}{c}{Step of the walk} \\
\cmidrule(lr){3-11}
Model & Measure & 0 & 1 & 2 & 3 & 5 & 8 & 10 & 12 & 15 \\
\midrule
\nfm   & Error   & 14.20 & 15.22 & 11.11 & 9.57 & 7.63 & 6.94 & 6.79 & --- & 6.72 \\
       & Regret  & 16.01 & 33.26 & 22.62 & 22.71 & 17.60 & 14.66 & 13.69 & 13.19 & 12.80 \\
       & Hedging & 24.9 & 91.5 & 58.1 & 66.6 & 42.5 & 30.9 & 28.9 & 28.1 & 28.6 \\
\nfmmh & Error   & 14.20 & 16.46 & 13.49 & 11.57 & 8.75 & 7.18 & 6.68 & --- & 6.27 \\
       & Regret  & 16.01 & 36.34 & 28.90 & 29.03 & 21.53 & 13.60 & 11.39 & 10.39 & 10.09 \\
       & Hedging & 24.9 & 95.9 & 87.5 & 89.7 & 67.0 & 44.5 & 35.2 & 28.5 & 22.6 \\
\bottomrule
\end{tabular}
\caption{The walk, step by step: mean location error (\% of the domain) and regret (\% of the range) on the \nScored{} in-family cases and hedging (\%) on all \nTies{} tie cases, of the walk's position after each step (point answers; step 0 is the start, the spline argmin; three seeds each; the harness prints the error at selected steps only, so step 12 is missing). The single-pass models give one answer: \st{} error 6.74, regret 15.70, hedging 22.7\%; \sttwo{} 6.28, 13.21 and 20.2\%.}
\label{tab:path}
\end{table}

\subsection{Tied minima: the form of the answer decides}
\label{sec:results-tie}

\begin{table}[t]
\centering
\footnotesize
\setlength{\tabcolsep}{3pt}
\begin{tabular}{>{\raggedright\arraybackslash}p{2.9cm}lrrrrr}
\toprule
& & \multicolumn{3}{c}{Exact tie (gap 0\%)} & \multicolumn{2}{c}{Hedging, gap} \\
\cmidrule(lr){3-5}\cmidrule(lr){6-7}
Form of the answer & Estimator & Hedging & In a floor & Capped regret & 2\% & 10\% \\
\midrule
\multirow{4}{2.9cm}{\raggedright One point, fitted to distance}
 & \nfm{} (walk)                       & 29.2\% & 36.5\% & 18.55 & 31.3\% & 25.2\% \\
 & \nfmmh, point (walk)                & 25.1\% & 44.0\% & 16.84 & 23.1\% & 19.7\% \\
 & \st{} (single pass)                 & 26.1\% & 45.3\% & 17.29 & 23.8\% & 18.1\% \\
 & \sttwo, point (single pass)         & 23.8\% & 52.5\% & 17.21 & 22.1\% & 14.8\% \\
\midrule
\multirow{3}{2.9cm}{\raggedright Minimum of a smoothed curve}
 & Spline argmin                       & 24.0\% & 40.8\% & 14.58 & 25.2\% & 24.0\% \\
 & Differential evolution on spline    & 20.0\% & 45.5\% & 11.63 & 21.5\% & 20.0\% \\
 & Basin-hopping on spline             & 26.0\% & 39.0\% & 15.78 & 29.0\% & 26.0\% \\
\midrule
\multirow{3}{2.9cm}{\raggedright Average or median of a belief over valleys}
 & \nfmmh, mixture mean                & 84.8\% & 4.6\% & 61.34 & 83.8\% & 70.8\% \\
 & \sttwo, mixture mean                & 81.0\% & 7.2\% & 57.09 & 79.6\% & 61.6\% \\
 & GP argmin posterior, median         & 10.5\% & 65.2\% & 7.59 & 13.2\% & 11.5\% \\
\midrule
Local fit, bandwidth by CV & Local quadratic (LOOCV) & 7.0\% & 73.0\% & 6.32 & 6.5\% & 7.5\% \\
\midrule
\multirow{5}{2.9cm}{\raggedright A mode, or the minimum of an unsmoothed curve}
 & \nfmmh, mode                        & 0.6\% & 96.7\% & 0.83 & 0.2\% & 0.1\% \\
 & \sttwo, mode                        & \textbf{0.3\%} & \textbf{97.6\%} & \textbf{0.67} & \textbf{0.0\%} & \textbf{0.0\%} \\
 & GP argmin posterior, mode           & 2.2\% & 84.0\% & 3.06 & 3.5\% & 2.0\% \\
 & GP posterior argmin                 & 1.2\% & 89.0\% & 1.55 & 1.8\% & 0.8\% \\
 & Lowest observation                  & 4.8\% & 63.7\% & 4.53 & 5.0\% & 5.0\% \\
\bottomrule
\end{tabular}
\caption{Behaviour at tied minima, grouped by the form of the answer: the harness's tie set, double wells whose two wells differ by 0, 2 or 10\% of the barrier, at four noise levels (400 cases per gap). Learned models: three seeds each. \emph{Hedging}: the answer is more than 5\% of the domain from every valley floor. \emph{In a floor}: within 2\% of one. \emph{Capped regret}: $f(\hat x)-f(x^{*})$ in percent of the barrier, with any value above one barrier counted as one barrier. The three answers of each two-head model come from one network. Paired differences over all tie cases, tie shapes resampled: \nfmmh{} $-$ \sttwo, point, hedging $+2.4$ [$-4.4$, $+9.4$] points, in a floor $-11.9$ [$-19.4$, $-4.6$]$^{*}$, capped regret $+0.91$ [$-3.68$, $+5.62$]; \nfm{} $-$ \st, hedging $+5.9$ [$-3.9$, $+18.4$]. Bold marks the best value in each column.}
\label{tab:tie}
\end{table}

When two valleys are equally deep, which one holds the global minimum is decided by differences far below the noise, and location error cannot score the answer: half the time the ``wrong'' valley is as good as the right one. What does matter is where the answer lies. The harness's tie set (100 double-well shapes, wells differing by 0, 2 or 10\% of the barrier, four noise levels; Table~\ref{tab:tie}) lets us ask this of every estimator at once.

\paragraph{One network, three answers.} The two-head models answer the question that a comparison between different models cannot: holding the network, its training and its data fixed, does the form of the answer decide? It does, in both designs. On the exact ties the \nfmmh's point answer hedges in 25.1\% of cases, its mode in 0.6\% and its mixture mean in 84.8\%; for the \sttwo{} the three are 23.8\%, 0.3\% and 81.0\%. The capped regret follows, from below 1\% of the barrier for the modes to 57 to 61\% for the mixture means. Between the designs, the difference is small beside these: the walk's point answer hedges as often as the single pass's ($+2.4$ [$-4.4$, $+9.4$] points over all tie cases), lies within 2\% of a floor less often ($-11.9$ [$-19.4$, $-4.6$]) and has the same capped regret ($+0.91$ [$-3.68$, $+5.62$]); the same holds with one head (exact ties: \nfm{} 29.2\%, \st{} 26.1\%; all tie cases: $+5.9$ [$-3.9$, $+18.4$] points). The modes of both designs commit in every cell.

\paragraph{The same grouping across all estimators.} Every estimator that returns one point fitted to its distance from the labelled minimum hedges in about a quarter of exact ties (the four learned point heads, 23.8 to 29.2\%), and so do estimators that take the minimum of a smoothed curve: the spline (24.0\%), and differential evolution (20.0\%) and basin-hopping (26.0\%), which search that spline. The median of the GP's posterior over the minimiser hedges in 10.5\% of cases. Estimators that name a mode, or take the minimum of a curve that keeps both wells, commit: the learned modes (0.3 and 0.6\%), the mode of the GP's argmin posterior (2.2\%), the GP's posterior-mean argmin (1.2\%) and the lowest observation (4.8\%); a local quadratic fit, whose cross-validated bandwidth is small enough to keep both wells, lies between (7.0\%). An average over the valleys hedges most: the learned mixture means, 81 to 85\%. The share of answers within 2\% of a floor, a second threshold, gives a similar grouping.

\paragraph{Mechanisms.} Three mechanisms are consistent with the pattern, and none is specific to neural networks. The first is the loss; this one we argue from the loss's form, since no model was trained with a different loss. At an exact tie the label is, from the data's point of view, a coin flip between two floors a distance $D$ apart. The point that minimises expected absolute error against such a label is not unique: every point between the floors is optimal, so a floor is no better than the midpoint. Under the Huber loss used to train the four point heads (threshold $\beta=0.05$), the expected loss at a floor exceeds that at an interior point by $\beta/4$ (Appendix~\ref{app:huber}): the interior is strictly preferred. A single point trained on distance therefore has no reason to commit, and a mild reason not to; that a network trained on distance averages between valid answers is the reason mixture density networks were introduced \citep{bishop1994mixture} and has been shown for multiple-hypothesis prediction \citep{rupprecht2017learning}, and what we add is its measurement across learned and classical estimators, its separation from averaging and smoothing, and its persistence where the data resolve the tie; its hedging is the optimal response to a loss that does not measure what a user pays, rather than a failure of the model. The second is averaging: the mean of a mixture with mass on two valleys lies between them by construction, whatever the network. The third is smoothing: a smoothing spline fitted to twenty samples, noisy or not (its smoothing has a floor, so even noiseless samples are smoothed), blurs two nearby wells into one broad valley whose minimum lies on the crest between them, and every optimiser that searches the spline inherits that minimum. Commitment comes from naming a mode, as the heaviest component of a mixture does and as the mode of the GP's argmin posterior does, or from taking the minimum of a curve that keeps both wells, as the GP's posterior mean and the lowest sample do. Two comparisons hold everything else fixed: the GP's median (10.5\%) and mode (2.2\%), read from one posterior, and the three answers of each two-head model, read from one network. In both the information is the same and the summary decides.

\paragraph{Where the data resolve the tie.} At an exact tie hedging is the loss-optimal response to a label that the data cannot resolve, so the question is whether it persists where the data do resolve it. Broken down by noise level (100 shapes per cell), it does for the point heads. With noiseless samples and wells that differ by 10\% of the barrier, where the GP posterior argmin and the learned modes never hedge, the point heads hedge in 10.7 (\sttwo) to 15.7\% (\nfm) of cases, and the mixture means in 43 to 53\%; at a 10\% gap the point heads hedge more in every noisy cell than in the noiseless one (\nfm{} 15.7\% at $\sigma/B=0$; 26.3, 31.3 and 27.7\% at $0.05$, $0.1$ and $0.2$). Part of the hedging of distance-trained point answers is therefore not explained by ambiguity in the data (Table~\ref{tab:hedgenoise}).

\paragraph{Griewank.} On Griewank, a function with adjacent valleys watched during development, the walk's point answers lie on a slope between valleys less often than the single pass's, with one head and with two (11.4 against 27.1\% and 9.6 against 21.3\%); this is exploratory, and Appendix~\ref{app:griewank} gives the details.

\subsection{Uncertainty}
\label{sec:results-calibration}

\begin{table}[t]
\centering
\small
\setlength{\tabcolsep}{3.2pt}
\begin{tabular}{lrrrrrrrr}
\toprule
Model / reading & $\bar s/\overline{|e|}$ & dev & PIT & ENCE & NLL & CRPS & $\rho(s,|e|)$ & E-AURC \\
\midrule
\nfm{}                                  & 1.13 & 0.045 & 0.044 & 0.115 & $-1.36$ & 5.25 & 0.60 & 1.43 \\
\nfmmh, point                           & \textbf{1.00} & 0.039 & 0.046 & 0.111 & $-1.51$ & 4.93 & 0.63 & 1.24 \\
\nfmmh, mixture mean                    & 1.02 & 0.062 & 0.039 & 0.130 & $-1.47$ & 4.92 & \textbf{0.80} & \textbf{0.89} \\
\nfmmh, full mixture                    & --- & --- & 0.037 & --- & \textbf{$-$2.18} & \textbf{4.42} & --- & --- \\
\st, Huber read-back                    & 1.08 & 0.041 & 0.043 & 0.105 & $-1.26$ & 5.15 & 0.54 & 1.58 \\
\st, as produced                        & 0.81 & 0.094 & 0.057 & 0.542 & $-1.13$ & 5.17 & 0.54 & 1.58 \\
\sttwo, point, Huber read-back          & 1.12 & 0.061 & 0.052 & 0.179 & $-1.39$ & 4.90 & 0.58 & 1.33 \\
\sttwo, point, as produced              & 0.83 & 0.042 & 0.033 & 0.362 & $-1.37$ & 4.87 & 0.58 & 1.33 \\
\sttwo, mixture mean                    & 1.05 & 0.055 & 0.028 & \textbf{0.092} & $-1.41$ & 4.92 & 0.76 & 0.97 \\
\sttwo, full mixture                    & --- & --- & 0.029 & --- & $-2.01$ & 4.44 & --- & --- \\
GP argmin posterior, median and spread  & 0.98 & \textbf{0.036} & 0.052 & 0.458 & $-0.96$ & 6.37 & 0.54 & 2.48 \\
GP argmin posterior, full               & --- & --- & \textbf{0.026} & --- & --- & 5.84 & --- & --- \\
\bottomrule
\end{tabular}

\caption{Per-case uncertainty on the \nScored{} in-family cases, every spread read as the scale $s$ of a Laplace distribution centred on the answer. Targets: $\bar s/\overline{|e|}=1$; coverage of $|e|\le s,2s,3s$ at $63.2$, $86.5$ and $95.0\%$, whose mean absolute deviation is \emph{dev}; PIT and ENCE 0. NLL is the Laplace negative log-likelihood; CRPS is in percent of the domain; E-AURC is the excess area under the risk--coverage curve, in percent of the domain. The point heads of the \st{} and the \sttwo{} were trained on the Huber value of their error; \emph{Huber read-back} converts them exactly to the Laplace scale, and is the reading used in every comparison. The spread of a \emph{mixture mean} is $\mathbb E|X-\bar x|$ under the mixture; the \emph{full mixture} is scored as a distribution (NLL, CRPS, PIT). The width of the heaviest component, the spread of a \emph{mode} answer, is about 0.3 of the error and is not a usable error bar (Section~\ref{sec:results-calibration}). The GP's spread is the mean distance of its 256 argmin draws from its answer, floored at 0.01 like the learned models' heads; the full posterior is sample-based, so it has CRPS and PIT but no NLL. The paired differences between the walk and the single pass are in Table~\ref{tab:calib-diff} (Appendix~\ref{app:calib-diff}).}
\label{tab:calibration}
\end{table}

Table~\ref{tab:calibration} scores every spread on the in-family cases. Three findings.

\emph{The walk's spread ranks its errors better, with one head and with two, and has the better likelihood under the read-back.} The \nfm's NLL is better than the \st's read back to the Laplace scale ($-1.36$ against $-1.26$), and the \nfmmh's better than the \sttwo's ($-1.51$ against $-1.39$; $-0.12$ [$-0.19$, $-0.06$], significant after the correction, Table~\ref{tab:primary}). With two heads the walk's spread also has a lower coverage deviation and a lower ENCE, and with one head and with two it follows the error more closely ($\rho$ 0.60 against 0.54, and 0.63 against 0.58). Keeping only the half of the answers it is most confident about, the \nfmmh's mean error is 2.01\% of the domain, against 2.32 for the \sttwo, 2.58 for the \nfm{} and 2.88 for the \st. The two kinds of point head were trained to different targets, the walk's on $|e|$ and the single pass's on the Huber value of its error, and the read-back to the Laplace scale is exact only if the errors are Laplace; part of the likelihood difference may therefore come from the training target rather than from the architecture, and the ranking measures ($\rho$, E-AURC), which do not depend on the read-back, are the cleaner comparison. On the hard levels and on the unseen functions the two-head models' point spreads are tied on every measure, walk against single pass; their full mixtures differ there in opposite directions (NLL $-0.06$ [$-0.11$, $-0.01$] on the hard levels, $+0.06$ [$+0.02$, $+0.11$] on the unseen functions).

\emph{Size.} The \nfm's raw spread is about 13\% too wide ($\bar s/\overline{|e|}=1.13$); the \nfmmh's has the right size (1.00), with coverage 72, 89 and 94\% at one, two and three spreads. One scale factor, fitted on half the cases and tested on the other half, removes the difference in size: after it the \nfmmh's NLL remains the best single spread ($-1.52$).

\emph{Distributions.} The full mixtures are the best distributions of all (NLL $-2.18$ for the \nfmmh{} and $-2.01$ for the \sttwo{}, $-0.17$ [$-0.21$, $-0.12$] walk minus single pass; CRPS 4.42 and 4.44, against 4.93 for the \nfmmh's point spread): a distribution with several modes describes a multi-valley problem better than any single spread can. The mixture means' spreads order the cases best of all spreads ($\rho$ 0.80 and 0.76; E-AURC 0.89 and 0.97), because the spread of a mean between valleys is large exactly when the answer is likely to be far from the minimum. The width of the heaviest component, the spread that goes with a mode answer, is about 0.3 of the error ($\bar s/\overline{|e|}=0.29$ and 0.31) and is not an error bar. The GP's full argmin posterior has the lowest PIT error among the full distributions (0.026), but its sample spread is a weak per-case uncertainty (E-AURC 2.48).

\subsection{The moving viewpoint}
\label{sec:results-ablations}

The training monitor's ablations (Appendix~\ref{app:ablations}) show that the summary vector and the updater carry the answer and that the learned bandwidth is unused. The updater re-reads all twenty samples at every step, and each sample token carries its offset $x_i-x_{t+1}$ from the walker's new position; the query also carries $x_{t+1}$. To see whether the model uses this moving viewpoint, we re-ran the three \nfm{} seeds on the \nScored{} in-family cases with parts of it held fixed and everything else, including the re-reads, unchanged (95\% intervals, seeds and cases resampled within cells). With the token offsets frozen at the start, $x_i-x_0$, the error rises by 31.3 [29.3, 33.4] points and regret by 28.1 [27.3, 29.0]; with the query's position frozen and the offsets left moving, nothing changes (error $-0.00$ [$-0.02$, $+0.01$], regret $-0.05$): the walker's position reaches the updater through the offsets. Frozen offsets lie outside what the model was trained on, so the size of that effect overstates the model's reliance. A small perturbation, the offsets and query taken one step behind ($x_t$ instead of $x_{t+1}$), raises the error by 0.47 [0.42, 0.53] and regret by 2.12 [1.82, 2.41]. The model therefore depends on where the walk now stands; the probe does not show that the walk's advantage over a single pass comes from this. We ran this probe on the \nfm{} only; whether a model trained without the offsets would do worse would require retraining.

\subsection{Matched computation: a looped Set Transformer}
\label{sec:results-looped}

\begin{table}[t]
\centering\footnotesize
\setlength{\tabcolsep}{3pt}
\begin{tabular}{lrrrrrrrr}
\toprule
& \multicolumn{4}{c}{In family} & \multicolumn{2}{c}{Hard levels} & \multicolumn{2}{c}{Unseen} \\
\cmidrule(lr){2-5}\cmidrule(lr){6-7}\cmidrule(lr){8-9}
Model & Error & Median & Regret & Right basin & Error & Regret & Error & Regret \\
\midrule
\nfm{} (walk) & \textbf{6.72} & \textbf{2.31} & \textbf{12.80} & 82.4\% & 13.67 & \textbf{25.07} & 9.29 & 23.08 \\
\st{} (single pass) & 6.74 & 2.84 & 15.70 & \textbf{82.8\%} & \textbf{13.41} & 27.96 & 9.57 & 25.56 \\
\stl{} (looped, 32 steps) & 7.38 & 2.87 & 15.12 & 81.2\% & 14.52 & 27.60 & \textbf{8.21} & \textbf{21.27} \\
\midrule
\multicolumn{9}{l}{Declared comparisons (Holm over the four)} \\
\multicolumn{9}{l}{\quad 1. Regret, \nfm{} $-$ \stl: $-2.32$ [$-2.94$, $-1.70$], Holm $p$ \textbf{$\le$0.0008}} \\
\multicolumn{9}{l}{\quad 2. Error, \nfm{} $-$ \stl: $-0.67$ [$-0.90$, $-0.44$], Holm $p$ \textbf{$\le$0.0008}} \\
\multicolumn{9}{l}{\quad 3. Regret, \stl{} $-$ \st: $-0.58$ [$-1.45$, $+0.23$], Holm $p$ 0.18} \\
\multicolumn{9}{l}{\quad 4. Error, \stl{} $-$ \st: $+0.65$ [$+0.45$, $+0.87$], Holm $p$ \textbf{$\le$0.0008}} \\
\bottomrule
\end{tabular}
\caption{The looped Set Transformer against the walk and the single pass, on the same cases as Table~\ref{tab:benchmark} (error and median error in percent of the domain, regret in percent of the range; three seeds each). The \stl{} matches the \nfm{} in data, training steps, parameters and operations per training step. The four comparisons were declared before any looped model was trained; bold in the comparisons marks significance after Holm's correction, and $\le$ marks a value at the bootstrap floor (5{,}000 draws). The \nfm{} and \st{} rows reproduce Table~\ref{tab:benchmark} exactly.}
\label{tab:looped}
\end{table}

The walk does more computation than a single pass, so its lower regret could come from computing more rather than from walking. The \stl{} tests this: the \st{} with a loop of 32 shared-weight steps that re-read the samples without moving (Section~\ref{sec:baselines}), matched to the \nfm{} in data, training steps, parameters and operations per training step. On one CPU it also takes as long to answer: 5.0~ms per case, against 5.2 for the \nfm{} and 1.4 for the \st. Its four comparisons were declared before it was trained (Table~\ref{tab:looped}). The \nfm's regret is lower than the \stl's by 2.32 [1.70, 2.94] points of the range and its error by 0.67 [0.44, 0.90]. Looping does not lower the single pass's regret significantly ($-0.58$ [$-1.45$, $+0.23$], Holm $p$ 0.18), and it raises its error by 0.65 [0.45, 0.87]. Giving the Set Transformer a loop with matched computation, which it settles in about eight steps, does not reproduce the walk's lower regret; it reproduces part of its mechanism, not the size of the effect (below). The moving-viewpoint probe shows that the trained \nfm{} depends on where it stands; only a walk retrained without the offsets could show that the advantage comes from it.

The rest is exploratory. The loop is in use: with its update frozen at its case average, the error rises from 7.3--7.5 to 37--59 points of the domain, and with no loop steps it is 50--58. But it settles early: after step 8, its answers move by 0.07 to 0.15 points of the domain on average over the remaining 24 steps, while the \nfm's answers move by 1.35 to 1.57 points between step 8 and step 15, the steps in which its regret falls (Section~\ref{sec:results-walk}). The loop does part of what the walk does: the \stl's answers in the right basin cost less than the \st's (regret 10.73 against 11.60; $-0.86$ [$-1.70$, $-0.06$]), about a third of the walk's gain there (9.03 against 11.60), while in the wrong basin its difference is not significant ($-1.40$ [$-3.64$, $+0.75$]); the \nfm's right-basin answers cost less than the \stl's by 1.70 [1.08, 2.31]. At the hard levels and under heavy-tailed noise, the \nfm's regret is again lower, by 2.53 [1.40, 3.69] and 2.63 [1.33, 3.90] points, and its spread has the better likelihood (NLL $-0.20$ [$-0.25$, $-0.15$], the \stl's read back from its Huber target). On the eleven unseen functions the order reverses: the \stl{} is the most accurate learned model, better than the \nfm{} on error (by 1.08 [0.38, 1.75]) and on regret (by 1.81 [0.27, 3.30]) and better than the \st{} (by 1.37 and 4.28), with cases resampled within each function and level. With the eleven functions resampled as wholes (computed after the run), the error difference to the \nfm{} holds (1.08 [0.15, 2.62]) but the regret difference does not (1.81 [$-0.82$, 5.42]); against the \st{} both hold (error 1.37 [0.26, 2.75], regret 4.28 [0.77, 9.39]). 

The \stl{} has weaknesses that a better-designed loop might not. The state entering the loop is not normalised (its spread across cases is of order $10^4$; the loop's layer normalisation rescales it at the first step), and the loop has no zero-initialised gate, one of the stabilisers the \nfm's updater needed (Section~\ref{sec:repair}). Its gradient guard skipped 1.3 to 3.6\% of training batches, against 0.6 to 0.9\% for the \nfm. Its point loss is the Set Transformer's, with no term on its loop steps, whereas the \nfm's loss supervises every step of the walk; an unsupervised loop may settle early for that reason. It was trained in one configuration and in about half the \nfm's wall-clock time (15.9 to 16.8 hours against 33 to 34). A better-conditioned looped Set Transformer might do better, and its error, 0.65 points worse than the \st's, may reflect this.

\subsection{Replication on fresh cases}
\label{sec:results-fresh}

\begin{table}[t]
\centering\footnotesize
\setlength{\tabcolsep}{3pt}
\begin{tabular}{llll}
\toprule
Comparison & Test cases (Tables~\ref{tab:primary}, \ref{tab:looped}) & Fresh cases & Holm $p$, fresh \\
\midrule
\multicolumn{4}{@{}l}{\textit{The $2\times2$ family (Holm over eight)}} \\
Regret, \nfmmh{} $-$ \sttwo, point & $-3.12$ [$-3.78$, $-2.49$] & $-2.94$ [$-3.62$, $-2.28$] & \textbf{$\le$0.0016} \\
Error, \nfmmh{} $-$ \sttwo, point & $-0.01$ [$-0.19$, $+0.17$] & $+0.04$ [$-0.13$, $+0.22$] & 1 \\
Regret, \nfmmh{} $-$ \sttwo, mode & $-1.45$ [$-1.93$, $-0.98$] & $-1.29$ [$-1.67$, $-0.92$] & \textbf{$\le$0.0016} \\
Error, \nfmmh{} $-$ \sttwo, mode & $-0.01$ [$-0.27$, $+0.25$] & $-0.15$ [$-0.40$, $+0.09$] & 0.84 \\
Tie hedging (points), point & $+2.4$ [$-4.4$, $+9.4$] & $+1.0$ [$-4.8$, $+6.5$] & 1 \\
NLL, point (read-back) & $-0.12$ [$-0.19$, $-0.06$] & $-0.11$ [$-0.16$, $-0.05$] & \textbf{0.01} \\
Regret, \nfm{} $-$ \st & $-2.90$ [$-3.76$, $-2.14$] & $-2.69$ [$-3.31$, $-2.09$] & \textbf{$\le$0.0016} \\
Error, \nfm{} $-$ \st & $-0.02$ [$-0.19$, $+0.13$] & $+0.09$ [$-0.14$, $+0.33$] & 1 \\
\midrule
\multicolumn{4}{@{}l}{\textit{The looped family (Holm over four)}} \\
Regret, \nfm{} $-$ \stl & $-2.32$ [$-2.94$, $-1.70$] & $-2.24$ [$-2.75$, $-1.72$] & \textbf{$\le$0.0008} \\
Error, \nfm{} $-$ \stl & $-0.67$ [$-0.90$, $-0.44$] & $-0.63$ [$-0.89$, $-0.35$] & \textbf{$\le$0.0008} \\
Regret, \stl{} $-$ \st & $-0.58$ [$-1.45$, $+0.23$] & $-0.45$ [$-1.09$, $+0.15$] & 0.15 \\
Error, \stl{} $-$ \st & $+0.65$ [$+0.45$, $+0.87$] & $+0.72$ [$+0.53$, $+0.92$] & \textbf{$\le$0.0008} \\
\bottomrule
\end{tabular}
\caption{The declared comparisons on the test cases and, once more, on a block of fresh cases (every case seed base moved by 50{,}000{,}000; case fingerprint \texttt{eefb730e3e3e}), with the same trained models, three seeds each (error in points of the domain, regret in points of the range, hedging in percentage points, NLL in nats; 95\% intervals; negative favours the first model). Bold: significant after Holm's correction within the family; $\le$ marks a value at the bootstrap floor. The plan for the fresh run was written before it (Section~\ref{sec:results-selection}).}
\label{tab:fresh}
\end{table}

Because the test cases were used more than once during the work (Section~\ref{sec:results-selection}), every declared comparison was run once more, with the same trained models, on a block of cases that no one had seen, drawn by the same generator and rules with every seed base moved (Table~\ref{tab:fresh}). All twelve comparisons keep their sign and their status after Holm's correction. The walk's regret is lower by 2.69 and 2.94 points of the range with one head and with two, and by 1.29 for the modes; its spread has the better likelihood; and the \nfm's regret is lower than the looped Set Transformer's by 2.24. The error comparisons of the $2\times2$ family remain not significant: on the fresh cases they are consistent with no difference larger than about 0.3 points of the domain with one head and 0.4 with two. Looping again raises the single pass's error and does not lower its regret significantly. On the fresh cases the \nfm{} reaches an error of 6.57 and a regret of 12.62, the \st{} 6.48 and 15.31.

\subsection{Further checks}
\label{sec:results-further}

Further analyses are in Appendix~\ref{app:further}. The walk's answer depends little on where it starts: from five starting points, the endpoints of a case lie within a median of 0.4 to 0.55\% of the domain of each other (Appendix~\ref{sec:results-path}). The models do not appear to exploit the tie between the noise and the well depth: with the noise tied to the function's range instead, the order of the models does not change; stronger classical baselines (the median and the mode of the GP's argmin posterior, and a local quadratic regression with a cross-validated bandwidth) do not change it either; and on the double-well family a Bayes rule that knows the prior and targets the basin chooses the right basin 7 to 12 points more often than the learned models at $\lambda=2$ (Appendix~\ref{sec:results-checks}). The median of the \nfm, \st{} and GP answers is more accurate in family than any one of them (Appendix~\ref{sec:results-ensemble}).

\subsection{Outside the training families}
\label{sec:results-oof}

On \nUnseen{} cases from eleven classical test functions \citep[e.g.][]{jamil2013literature} that belong to none of the thirteen families and were never used during development, each seen through a random window of 60 to 100\% of its native domain, randomly mirrored, with $y$ in the function's native units (Appendix~\ref{app:oof}), the learned models lose their advantage. The GP is the most accurate estimator (posterior median 7.55, posterior argmin 7.85), ahead of the \sttwo{} (9.07), the \nfmmh{} (9.25), the \nfm{} (9.29; seeds 8.67, 9.37 and 9.82) and the \st{} (9.57), and it chooses the right basin more often (80.8\% against 72.9 to 75.0\%). With the eleven functions resampled, the GP posterior median's lead over the \nfm{} is significant ($-1.74$ [$-3.48$, $-0.29$]) and the GP's regret is far lower (13.0 against 23.1); the four learned models are tied on error, and the walk's lower regret survives the resampling of whole functions with one head (\nfm{} $-$ \st{} $-2.48$ [$-4.92$, $-0.43$]) but not with two (\nfmmh{} $-$ \sttwo{} $-1.27$ [$-3.50$, $+0.82$]; whole functions resampled, 5{,}000 draws). The looped Set Transformer of Section~\ref{sec:results-looped} is more accurate here than all four (8.21; exploratory) and has lower regret than the \nfm{} (21.27 against 23.08); with whole functions resampled its error lead over the \nfm{} holds (1.08 [0.15, 2.62]) but its regret lead does not (1.81 [$-0.82$, 5.42]): against it, the walk's in-family advantage does not carry over to these functions, so the inductive bias we hoped the walk would carry is not visible outside the training families. The GP's lead is largest at low noise (at $\lambda=0$, 3.28 against 6.11 for the \nfm) and gone at $\lambda=1$ for the posterior argmin (14.61 against 13.89), though not for the posterior median (13.41). Because the learned models' inputs separate the shape of the samples from their scale, the deficit is not a matter of the units of $y$ alone; the shapes of these functions differ from those of the training families. On the five functions monitored during development (\nMonitored{} cases; not independent evidence) the picture is the same and sharper: GP posterior median 5.18, against 10.27 to 10.90 for the learned point answers, with every learned answer at 27 to 32\% error on Schwefel. We state this as a limitation: the learned estimators carry no guarantee outside the families they were trained on.

\subsection{Cost}
\label{sec:results-cost}

The learned models have nearly the same size (1.33M parameters). On a two-core CPU the walk models take 2.5~ms per case and the Set Transformers 0.75 to 0.78~ms, plus about 1.0~ms to build the inputs from the samples (the same inputs for all four); the GP with its 256 posterior draws takes 62~ms per case, and differential evolution with basin-hopping 20~ms. A walk is 3.2 to 3.3 times slower than a single pass in the network and 2.0 times slower with the inputs included; it is about 17 times faster than the GP. The mixture head adds no measurable time. The \stl, timed in a separate run on another CPU, takes 5.0~ms per case against 5.2 for the \nfm{} and 1.4 for the \st{} on that machine. These ratios depend on the CPU and on the GP implementation (scikit-learn, 256 posterior draws). Training is a different matter: a run of a walk model took about ten times as long as a run of a Set Transformer (Section~\ref{sec:model}).

\section{Discussion}
\label{sec:discussion}

The $2\times2$ design separates three effects that a comparison of two different models cannot. Matched learned models reach the same location error whether they walk or answer in one pass, with one head or two; the \nfm's answers have lower regret, mostly because those in the right basin lie closer to the floor, and the walk earns this in its second half, after its error has settled (Section~\ref{sec:results-walk}). The mixture loss lowers the error by about the same amount in both designs. Where two valleys are nearly equal, the form of the answer decides more than the computation or the loss.

What causes the regret effect is narrowed but not settled. The \nfm{} differs from a Set Transformer in its encoder, local read, activation, trajectory loss and training time, not only in walking. A loop of matched computation, which settles in about eight steps, does not reproduce the effect within the training families (Section~\ref{sec:results-looped}), and the moving-viewpoint probe shows that the trained \nfm{} depends on where it stands; a walk retrained without the offsets, and walks trained for fewer steps, are the natural next tests. On the unseen functions the looped model has the lower error (exploratory); worse inside the training families and better outside them is also what a model that fits its training families less tightly would show, so this need not mean that looping carries a better inductive bias. What no design changes is the pull of a distance loss towards the middle of two equal valleys: the walk's point answer hedges as often as the single pass's on the constructed ties, although with two heads it lies on a floor less often (by 11.9 points).

A user who cares about regret should choose the form of the answer before the architecture. The modes of the two-head models have lower regret than their point answers (6.53 and 7.98); the \nfmmh's mode has the lowest regret of any learned answer, and the walk lowers the mode's regret by 1.45 (significant after the correction), while for the single pass the mixture-only Set Transformer of Appendix~\ref{app:families-declared} is lower than the \sttwo's mode (7.52); the GP's posterior argmin, which also lies on a valley floor, is lower still. A user who wants a distribution should take the full mixture, which is the best distribution in this paper, and should not summarise it by its mean.

We do not propose the walk as a replacement for attention-based set models, which, given the same inputs, match it on location error with no task-specific architecture. Its updater is itself an attention read over the samples; what the iterative design adds is that the read is repeated from a moving position, with the samples described relative to it, at about twice the time per answer, inputs included (Section~\ref{sec:results-cost}). The designs are complementary rather than competing: the median of the \nfm, the \st{} and the GP is as accurate on the mean as the two-head models.

The iterative design has a failure mode the single-pass design does not: its update module can saturate and leave a plausible single-pass network behind (Section~\ref{sec:repair}). The check that detects it, replacing a module's output by a constant, is cheap and decisive, and we recommend it for any model with a large recurrent module.

\paragraph{Perspective.} This work began from a view of a learned model as an algorithm, with a state revised step by step and a read of the data from a position that moves, in the tradition of unrolled iterative methods. The evidence supports part of that view: the walk can be followed and probed step by step, a module that had silently stopped working was found because its parts can be switched off, and within the training families the \nfm's answers have lower regret than a single pass's, looped or not. It is not more accurate, and outside the training families the hoped-for inductive bias is not visible; whether the design pays off on harder problems, in higher dimensions, or with an answer that names a mode is what we would like to test next.

\paragraph{Limitations.} The walk is not more accurate than the single pass on mean location error, in any cell of the design. On regret the GP posterior argmin is better than every learned answer, and outside the training families the GP is the most accurate estimator; there is no guarantee for the learned models outside the families they were trained on. On the double-well family, a Bayes rule that knows the prior and targets the basin chooses the right basin 7 to 12 points more often than the learned models at $\lambda=2$ (Appendix~\ref{sec:results-checks}). Each learned model has three training seeds; the intervals include that variation, but three seeds estimate it poorly, and intervals between models may be too narrow. The Set Transformers are matched to the walk models in parameters, inputs and the mixture head and its loss, not in their point and spread losses (Section~\ref{sec:baselines}) or in compute: a walk trains about ten times as long and answers about twice as slowly, inputs included, and the \nfm's design was chosen among variants while each Set Transformer had one configuration. The looped Set Transformer matches the walk's operations per step and its time per answer, but not its wall-clock training time, it is less well conditioned than the walk, and its loss has no term on its loop steps, whereas the walk's supervises every step (Section~\ref{sec:results-looped}); a better loop might do better. It was added last, and a quick run on partly overlapping cases was seen before its standard run. The two-head models were added to an existing comparison; the first seed of the \nfmmh{} was scored on the test cases several times during its training, and the $2\times2$ family was declared after the first seeds of both two-head models had been scored (Section~\ref{sec:results-selection}). The point spreads of the walk and the single pass were trained to different targets, so the likelihood comparison carries an assumption that the ranking comparison does not. The tie observation is made on synthetic double wells and one canonical function, and its threshold (5\% of the domain from every floor) is a convention; we have not tested it on real data. Eleven unseen functions give wide intervals. The five monitored functions were watched during development and are not independent. Our Bayes reference covers one family and only the two noisiest levels, and it is an approximation built from four million prior draws. The learned bandwidth multiplier of the local read is unused. The activation was not ablated, and the Set Transformers use standard activations. The benchmark accepts a sample set only if at least one sample lies 0.02 of the domain or more on each side of the true minimiser, so the minimum is never outside the sampled range; a learned model trained under the same rule can exploit it, while the classical estimators search between the first and last sample. Regret is normalised by the function's range, which weights families with steep edges less. The \nfm's design was chosen among variants partly on test results (Section~\ref{sec:results-selection}); the declared comparisons were therefore replicated on fresh cases, with the same trained models (Section~\ref{sec:results-fresh}). No model was trained to the end of its learning-rate schedule: at step \trainEpochs{} the rate is still about $1.9\times10^{-4}$, and an annealed Set Transformer might close part of the gap. The number of samples ($n=20$) and of walk steps ($T=15$) are fixed throughout. The problem is one-dimensional; whether any of these findings carries over to higher dimensions is untested.

\section{Conclusion}

We built a benchmark and a self-checking harness for estimating the global minimum of a noisy one-dimensional function from twenty samples, and used them for a controlled comparison of an iterative neural estimator that we introduce with a same-size Set Transformer, each with one output head and with a second, mixture head, given the same inputs and, with two heads, the same mixture loss, and with classical estimators, on identical cases, over three training seeds each, with paired statistics and declared comparisons, reporting what the comparison does and does not support. Within the training families the walk and the single pass are tied on location error, with one head and with two; the \nfm{} has lower regret, by about the same amount with either head, as well as a lower median error and a spread that ranks its errors better, and within the training families a Set Transformer looped to the walk's computation does not reproduce it; and the mixture loss lowers the error by about the same amount in both. At nearly tied minima the form of the answer, read from one network, decides whether an estimator commits to a valley: point answers trained on distance hedge, even when the data resolve the tie, modes commit, and mixture means hedge most. The GP posterior argmin has the lowest regret, and outside the training families it is the better estimator.

\bibliographystyle{plainnat}
\paragraph{Use of AI assistants.} We used an AI assistant (Claude, Anthropic) in this work: for drafting and editing the text, for writing, reviewing and debugging code, including the evaluation harness and the analysis scripts, and for checking reported numbers against the run files. The ideas, the experiments and the conclusions are the authors', and the authors checked every claim.

\bibliography{refs}

\clearpage
\appendix
\section{Benchmark specification}
\label{app:bench}

This appendix specifies every case set in the paper completely enough to regenerate it without the released code. Every statement is read from the code that produced the reported numbers: the training sources (\texttt{train\_source.py} in each run folder, whose generator, sampler, noise and input builder the harness imports rather than reimplements), and the released test script \texttt{NFM\_tests\_two\_heads\_standard.py}, whose sections are the benchmark harness, the Griewank tie test, the noise and Bayes checks and the ensemble.

\subsection{Frame, grids and conventions}
\label{app:frame}

Every function is written on $x\in[0,1]$. All dense grids are uniform on $[0.001,0.999]$; we write $G_N$ for the grid of $N$ points. The generator uses $G_{3000}$ to draw functions, locate the true minimiser and compute the well depth; the harness uses $G_{2000}$ for basins and regret, $G_{1200}$ for the GP posterior-mean argmin and $G_{1000}$ inside the input builder. Function values are not rescaled: the twenty clean values $f(x_i)$ are shifted by their own minimum, $y^{\circ}_i=f(x_i)-\min_j f(x_j)$, and the noise is added to $y^{\circ}_i$ (the $\tilde y$ of the model inputs, Appendix~\ref{app:inputs}, is the noisy $y$ rescaled to $[0,1]$). Errors are $|\xhat-\xs|$ in percent of the domain. Every case is identified by an integer seed (Appendix~\ref{app:casesets}); the harness saves and restores NumPy's global random state around every draw, so that building one set cannot shift another, and the classical estimators' own randomness is seeded from the case.

\subsection{The thirteen function families}
\label{app:families}

A function is drawn by choosing a family (with the mixture weights below in training; fixed per cell in evaluation), then a centre $m\sim U(0.05,0.95)$, then the family's parameters, in the order they are listed. $U(a,b)$ is uniform on the interval, $U\{a,\dots,b\}$ uniform on the integers, and all draws are independent.

\begin{description}\itemsep2pt
\item[Bowl (0.09).] $f(x)=s\,(x-m)^{p}$; $p\in U\{2,4,6,8\}$, $s\sim U(1,8)$. One basin; $\xs=m$.
\item[Polynomial (0.08).] $f(x)=\pm20\prod_{j=1}^{D}(x-r_j)$; $D\sim U\{3,\dots,8\}$, $r_j\sim U(-0.2,1.2)$; the sign is $+$ for even $D$ and $-$ for odd $D$. The roots may lie outside $[0,1]$, so the number of wells inside the domain varies.
\item[Periodic (0.09).] $f(x)=b(x-m)^2+\sum_{j=1}^{J}a_j\sin\!\big(f_j\pi x(1+\kappa x)+\phi_j\big)$; $J\sim U\{2,3,4\}$, then the vectors $f_j\sim U(2,9)$, $a_j\sim U(0.2,0.9)$, $\phi_j\sim U(0,2\pi)$; the chirp $\kappa$ is $U(0,3)$ with probability $\tfrac12$ and $0$ otherwise; $b\sim U(2,6)$.
\item[Regular periodic (0.09).] $f(x)=b(x-m)^2+A\big(1-\cos 2\pi F(x-m)\big)$; $F\sim U(3,8)$, $A\sim U(0.5,2)$, $b\sim U(3,10)$. Both terms vanish at $m$, so $\xs=m$; the neighbouring valleys are shallower by the bowl term only.
\item[Double well (0.10).] $f(x)=a\big((x-x_1)(x-x_2)\big)^2+t\,(x-x_1)$; separation $\Delta\sim U(0.25,0.70)$, $x_1\sim U(0.06,\,0.94-\Delta)$, $x_2=x_1+\Delta$, barrier $B\sim U(0.5,4)$, $a=16B/\Delta^4$ (the untilted barrier between the wells has height $B$), and tilt $t=\pm gB/\Delta$ with $g\sim U(0.005,0.22)$ and a sign that is $-$ with probability $\tfrac12$. The two wells differ in depth by about $gB$, i.e.\ by 0.5\% to 22\% of the barrier.
\item[Multi-well (0.08).] $f(x)=b(x-0.5)^2-\sum_{j=1}^{K}D_j\exp\!\big(-((x-\ell_j)/w_j)^2\big)$; $K\sim U\{2,3,4\}$, $\ell_1=m$ and $\ell_j\sim U(0.08,0.92)$ for $j\ge2$, $D_1\sim U(1.2,2.5)$, $D_j=D_1(1-g_j)$ with $g_j\sim U(0.005,0.22)$ for $j\ge2$, widths $w_j\sim U(0.065,0.115)$, $b\sim U(0.02,0.2)$.
\item[Spline (0.06).] $f(x)=5\,S(x)$, where $S$ is the cubic B-spline with knot vector $(0,0,0,0,0.5,1,1,1,1)$ and five coefficients drawn from $U(-1,1)$.
\item[Deceptive (0.06).] $f(x)=4(x-M)^2-D\exp\!\big(-((x-m)/w)^2\big)$; $w\sim U(0.06,0.13)$, $D\sim U(3,6)$. The bowl's centre $M$ is placed away from the narrow well: $M=1-m$ if $|1-2m|>0.25$, otherwise $M=m+0.45$ for $m<0.5$ and $M=m-0.45$ for $m\ge0.5$. The broad bowl is the decoy; the narrow well at $m$ is usually the global minimum, which is always located numerically.
\item[Growing ripple (0.09).] $f(x)=-\big(1+G(u-0.5)\big)\cos 2\pi F(x-m)+b(x-m)^2$; $F\sim U(2.5,6)$, a phase $\sim U(0,2\pi)$ that is drawn and not used (it matters only for reproducing the random stream), $G\sim U(0.6,1.55)$, orientation $u=1-x$ with probability $\tfrac12$ and $u=x$ otherwise, $b\sim U(0.2,1.1)$.
\item[Decaying ripple (0.06).] $f(x)=-\cos\!\big(2\pi F(x-c)\big)/\big(1+k|x-c|\big)+b(x-c)^2$; $c\sim U(0.15,0.85)$, $F\sim U(3,8)$, $k\sim U(0.6,3.5)$, $b\sim U(0.1,1)$. The centre $m$ is drawn but not used by this family.
\item[Exponential envelope (0.06).] $f(x)=-e^{-\theta|x-m|}-A\cos 2\pi F(x-m)$; $\theta\sim U(1.5,6)$, $A\sim U(0.35,0.95)$, $F\sim U(3,9)$. $\xs=m$.
\item[Plateau with valleys (0.06).] $f(x)=-\big|\sin(\pi Fx+\phi)\big|^{P}+b(x-m)^2$; $P\in U\{6,10,16\}$, $F\sim U(2,4.5)$, $\phi\sim U(0,\pi)$, $b\sim U(0.1,0.4)$. The high power flattens the curve between narrow, nearly equal valleys; the weak bowl decides between them.
\item[Spectral (0.08).] $f(x)=b(x-m)^2+\tau x+A\sum_{k=1}^{K}c_k\cos(2\pi kx+\phi_k)$, with $\xi\sim U(1,3)$, $K\sim U\{8,\dots,24\}$, $c_k=k^{-\xi/2}\big/\big(\tfrac12\sum_{k'=1}^{K}k'^{-\xi}\big)^{1/2}$ so that the fluctuation has unit variance over random phases, $\phi_k\sim U(0,2\pi)$, $A\sim U(0.08,0.55)$, $b\sim U(2,8)$, $\tau\sim U(-0.5,0.5)$. The power spectrum of the fluctuation falls as $k^{-\xi}$.
\end{description}

\paragraph{Acceptance tests.} A drawn function is evaluated on $G_{3000}$ and accepted only if (i) the evaluation succeeds and every value is finite; (ii) its range $R=\max f-\min f$ on the grid lies in $[10^{-4},10^{6}]$; (iii) $\xs\in[0.05,0.95]$; (iv) the well depth $d$ of Appendix~\ref{app:noise} is finite and exceeds $10^{-9}$; and (v) the global basin is not too narrow: the fraction $w_b$ of grid points with $f\le\min f+0.1R$ is at least $0.03$. A rejected draw is replaced by a new draw from the same family (new centre and parameters, continuing the random stream); after 25 consecutive rejections the generator gives up and the case is skipped. No case of any evaluation set was skipped. Table~\ref{tab:famstats} gives the measured properties of every family.

\begin{table}[t]
\centering\small
\setlength{\tabcolsep}{3.5pt}
\resizebox{\linewidth}{!}{%
\begin{tabular}{lrrrrrrrrr}
\toprule
Family & Weight & Basins & Multi-basin & Gap & $d/R$, median [10\%, 90\%] & Range $R$ & $w_b$ & Resolvable & Tie cases \\
\midrule
Bowl                 & 0.09 & 1 (1) & 0.0\% & --- & 1.34 [0.00, 42.9] & 0.98 & 0.706 & 100.0\% & 0 \\
Polynomial           & 0.08 & 2 (4) & 97.9\% & 16.5 & 15.6 [0.27, 93.4] & 0.94 & 0.254 & 100.0\% & 28 \\
Periodic             & 0.09 & 5 (17) & 100.0\% & 7.2 & 7.24 [1.23, 24.9] & 3.65 & 0.097 & 95.1\% & 8 \\
Regular periodic     & 0.09 & 6 (9) & 100.0\% & 4.3 & 4.25 [1.92, 10.1] & 5.00 & 0.127 & 99.3\% & 0 \\
Double well          & 0.10 & 2 (2) & 100.0\% & 0.3 & 0.32 [0.03, 3.5] & 46.02 & 0.693 & 100.0\% & 99 \\
Multi-well           & 0.08 & 2 (4) & 86.8\% & 32.3 & 32.3 [2.85, 99.8] & 2.39 & 0.071 & 95.8\% & 5 \\
Spline               & 0.06 & 2 (3) & 59.7\% & 74.9 & 46.8 [6.79, 87.6] & 4.98 & 0.253 & 100.0\% & 1 \\
Deceptive            & 0.06 & 2 (2) & 95.1\% & 56.3 & 56.3 [42.66, 70.3] & 5.65 & 0.071 & 96.5\% & 0 \\
Growing ripple       & 0.09 & 5 (7) & 100.0\% & 6.4 & 6.40 [0.84, 13.6] & 2.74 & 0.075 & 96.5\% & 18 \\
Decaying ripple      & 0.06 & 6 (9) & 100.0\% & 14.8 & 14.8 [6.27, 26.2] & 1.87 & 0.046 & 93.8\% & 0 \\
Exponential envelope & 0.06 & 6 (10) & 100.0\% & 22.9 & 22.9 [12.49, 33.2] & 2.18 & 0.042 & 92.4\% & 0 \\
Plateau with valleys & 0.06 & 4 (6) & 100.0\% & 1.3 & 1.31 [0.22, 4.3] & 1.09 & 0.078 & 93.8\% & 62 \\
Spectral             & 0.08 & 8 (18) & 98.6\% & 4.1 & 4.08 [0.59, 16.4] & 2.88 & 0.174 & 98.6\% & 29 \\
\midrule
All (1{,}872 draws) & 1.00 & 4 & 87.6\% & & & & & 97.1\% & 250 \\
\bottomrule
\end{tabular}}
\caption{Properties of the thirteen families, measured on 1{,}872 draws from the training generator (144 per family, 24 at each noise level). \emph{Basins}: median (maximum) number of watershed basins of $f$ on $G_{3000}$. \emph{Multi-basin}: share of cases with more than one basin. \emph{Gap}: median height of the runner-up basin's floor above the global minimum, in percent of $R$ (undefined for a single basin). \emph{$d/R$}: well depth in percent of the range, median and 10th and 90th percentiles; the median differs by a factor of about 176 between double wells and deceptive functions, which is why the noise is scaled to $d$ and not to $R$. \emph{Range $R$}: median, in the function's own units. \emph{$w_b$}: median fraction of the grid within 10\% of the range above the minimum, in any basin. \emph{Resolvable}: a sample lies within $\max(w_b,0.06)$ of $\xs$. \emph{Tie cases}: gap below 1\%.}
\label{tab:famstats}
\end{table}

\subsection{Sample locations}
\label{app:sampler}

The twenty locations of a case are drawn as follows.
\begin{enumerate}\itemsep1pt
\item Choose a sampler: uniform, Gaussian or Beta, with probabilities $0.50$, $0.25$, $0.25$.
\item Draw twenty points: uniform, $x_i\sim U(0.001,0.999)$; Gaussian, a centre $c\sim U(0.1,0.9)$ and a width $s\sim U(0.10,0.30)$, then $x_i\sim\mathcal N(c,s^2)$; Beta, $a\sim U(0.5,3)$ and $b\sim U(0.5,3)$, then $x_i\sim\mathrm{Beta}(a,b)$.
\item Clip to $[0.001,0.999]$ and sort.
\item Enforce a minimum spacing of $0.005$: for $i=2,\dots,20$, if $x_i-x_{i-1}<0.005$ set $x_i=x_{i-1}+0.005$. If this pushes $x_{20}$ past $0.999$, set $x_{20}=0.999$ and run the same rule backwards ($x_i=x_{i+1}-0.005$ where the spacing is violated, $i=19,\dots,1$), then clip to $[0.001,0.999]$ again.
\item Accept if the twenty points are distinct and bracket the minimum, $\min_i x_i\le\xs-0.02$ and $\max_i x_i\ge\xs+0.02$. Otherwise return to step 1, choosing the sampler anew.
\item After 400 rejections, draw once more and stretch the draw affinely onto $[0.001,0.999]$, keeping its relative spacing. This fallback was never needed on the reported cases.
\end{enumerate}
Because the Gaussian and Beta samplers miss the minimum more often, rejection shifts the mixture: of the \nScored{} in-family evaluation cases, 3{,}025 (58.2\%) are uniform, 1{,}057 (20.3\%) Gaussian and 1{,}118 (21.5\%) Beta. The same shift happens in training, since the harness calls the training sampler. The rollout of the \nfm{} starts at the spline argmin (Appendix~\ref{app:inputs}) or at a point drawn from $U(0.05,0.95)$, with probability $\tfrac12$ each in training; in every case set reported here the rollout starts at the spline argmin. A case is called \emph{resolvable} if some sample lies within $\max(w_b,0.06)$ of $\xs$; training down-weights unresolvable cases to $0.3$.

\subsection{Well depth and noise}
\label{app:noise}

Let $y_1,\dots,y_N$ be the values of $f$ on $G_{3000}$ and $k^{*}$ the index of the global minimum. The candidate rivals are the interior local minima ($y_k\le y_{k-1}$ and $y_k<y_{k+1}$), together with the first grid point if $y_1<y_2$ and the last if $y_N<y_{N-1}$, i.e.\ an endpoint towards which the curve descends. Excluding $k^{*}$, the well depth is
\[
d=\min_{k\ \text{rival}}y_k-y_{k^{*}},\qquad\text{or}\qquad d=\min(y_1,y_N)-y_{k^{*}}\ \text{if there is no rival.}
\]
The observations are $y_i=\tilde y_i+\sigma\varepsilon_i$ with $\sigma=\lambda d$ and $\varepsilon_i$ independent standard normal (Appendix~\ref{app:casesets} gives the two non-Gaussian variants). In training $\lambda$ is drawn uniformly from $\{0,0.25,0.5,1,2,4\}$ with no jitter; in evaluation every (family, level) cell has a fixed $\lambda$. $\lambda=0$ cases are noiseless and draw no noise. The Griewank test computes $d$ on $G_{2000}$ with one extra condition, that a candidate rival counts only if the curve rises somewhere between it and the global minimum; the condition matters only when the minimum falls between two grid points of equal value. In the tie set the noise is set as a fraction of the barrier instead (Appendix~\ref{app:casesets}).

\subsection{Inputs derived from the samples}
\label{app:inputs}

The four learned models receive the same inputs, built by the same code: four per-sample channels $(x_i,\tilde y_i,y'_i,c_i)_{i=1}^{20}$, a resolution radius $\delta$ and five descriptors $\mathbf g$, all computed from the sorted noisy samples as follows after the normalisation described at the end of this subsection, which also adds the fifth descriptor; the \nfm{} also receives a start $x_0$; the classical estimators receive $(x_i,y_i)$ only. The steps below are written for samples $y$; the input builder applies them to the normalised $\tilde y$. Write $\Delta y$ for the 19 successive differences of $y$ in the order of $x$, $\Delta^2y$ for the 18 second differences, and $\mathrm{MAD}(v)=\mathrm{median}|v-\mathrm{median}(v)|$.
\begin{enumerate}\itemsep1pt
\item Noise variance, $\hat v=\big(\mathrm{MAD}(\Delta y)/0.6745\big)^2/2$.
\item Smoothing spline: SciPy \texttt{splrep}$(x,y,k=3,s=S)$ with $S=0$ if $\hat v<10^{-10}$ (interpolation) and $S=\max(20\hat v,\,20\times10^{-4})$ otherwise; if the fit fails, it is repeated with $S=20\times10^{-4}$, and if that fails too the case is discarded and redrawn (no reported case was). With $S=0$ the spline interpolates the samples; this path is almost never taken, because for a smooth function the differences of noiseless samples do not vanish and $\hat v$ is above the threshold, so noiseless samples are smoothed with the floor $S$ (and $\hat v$ then partly measures the function's own curvature). For a smooth function cubic spline interpolation would be fourth-order accurate in the sample spacing and its slope third-order: the classical bounds are proved for the complete spline \citep{birkhoff1964error, hall1968error}, and the not-a-knot spline that SciPy computes has the same orders \citep{deboor2001practical}; with $S>0$ it is a smoothing spline \citep{wahba1990spline}, which trades this accuracy for robustness to the noise.
\item Slopes, $y'_i=S'(x_i)$, clipped to $[-25,25]$ (non-finite values set to 0 or $\pm25$).
\item Coefficients: the spline's B-spline coefficient vector, padded with zeros or truncated to length 20 and clipped to $[-5,5]$ (non-finite values set to 0 or $\pm1$); $c_i$ is its $i$-th entry.
\item Spline argmin $x_s$: the argmin of the spline on $G_{1000}$ restricted to $[\min x_i,\max x_i]$. This is the rollout start $x_0$. The ``spline argmin'' baseline applies the same routine to the raw samples rather than the normalised ones; because the smoothing floor below is absolute, the two splines, and their argmins, can differ (by more than $10^{-3}$ of the domain in about 7\% of noiseless cases).
\item Noise scales: $d_1=\mathrm{MAD}(\Delta y)/(0.6745\sqrt2)$ and $d_2=\mathrm{MAD}(\Delta^2y)/(0.6745\sqrt6)$, which agree for white noise; $\hat\sigma=\max(d_1,10^{-6})$.
\item Well count and separability: $n_w$ is the number of interior local minima of the spline on $G_{1000}$ (at least 1), $m_{\mathrm{eff}}=\max(20/n_w,1)$; the gap $\Gamma$ is the lowest spline value more than $0.15$ from $x_s$ minus the value at $x_s$ (floored at 0), and $z=\Gamma/\big(\hat\sigma\sqrt{2/m_{\mathrm{eff}}}\big)$.
\item Resolution radius: with $f''$ the central-difference second derivative of the spline at $x_s$ on $G_{1000}$, the radius is $\delta=\big(2\hat\sigma/(\sqrt{m_{\mathrm{eff}}}\,f'')\big)^{1/2}$ if $f''>10^{-9}$ and $\delta=0.25$ otherwise, clipped to $[10^{-3},0.25]$.
\item Descriptors (noise scale, resolution, separability, whiteness):
\[
\begin{aligned}
g_1&=\mathrm{clip}_{[-4,4]}\big(\ln(\hat\sigma+10^{-3})+3\big), & g_2&=\mathrm{clip}_{[-4,4]}\big(\ln(\delta+10^{-3})+3\big),\\
g_3&=\tanh(z/4), & g_4&=\mathrm{clip}_{[0,3]}\big(d_1/\max(d_2,10^{-6})\big)-1.
\end{aligned}
\]
\end{enumerate}
The \st{} and the \sttwo{} receive the four per-sample channels as tokens and the five descriptors $\mathbf g$ and $\delta$ as globals; they do not use $x_0$.

\paragraph{Scale-free inputs.} With $D_y=\max_iy_i-\min_iy_i$ of the twenty raw observations ($D_y=1$ if $D_y<10^{-12}$), the input builder of the learned models replaces $y$ by $\tilde y=(y-\min_jy_j)/D_y$ and computes steps 1 to 9 from $\tilde y$, so that the spline, $y'$, $c$, the noise scales, $z$, $\delta$ and $x_0$ are those of the normalised samples; $g_1$ becomes the noise scale relative to $D_y$, and a fifth descriptor $g_5=\mathrm{clip}_{[-8,8]}(\ln D_y)$ carries the scale (over 6{,}000 training draws $\ln D_y$ ranges from $-6.96$ to $+7.89$, median $+1.10$). Multiplying the observations by $a>0$ changes $g_5$ by $\ln a$ and nothing else; adding a constant changes nothing.

\subsection{Basins, gap and the tie-aware measures}
\label{app:metrics}

\paragraph{Basins.} The harness computes the basins of the clean function on $G_{2000}$. The minima are the interior local minima ($y_k\le y_{k-1}$ and $y_k<y_{k+1}$) and each endpoint towards which the curve descends; if there is none, the global argmin. Consecutive minima are separated at the highest grid point between them (watershed basins). An answer is assigned the basin of the grid point at its location, and it is in the right basin if that is the basin of $\xs$.

\paragraph{Regret.} $\big(f(\xhat)-f(\xs)\big)/R$ in percent, with $R$ the range of $f$ on the grid and $f$ the clean function. On the tie set regret is measured in percent of the barrier $B$ between the wells, and \emph{capped regret} counts any value above $B$ as $B$; the harness also reports the share of answers within 2\% of a domain edge (at most 3.2\% for any estimator and 0.1\% for any learned model).

\paragraph{Tie measures.} On the tie set the floors are the two wells $x_1,x_2$ of the construction. An answer \emph{hedges} if it is more than $0.05$ from both floors and is \emph{in a floor} if it is within $0.02$ of one; for gaps above zero, \emph{right valley} is the share of answers in the basin of the deeper well. On Griewank, the floors $u_b$ and ridges are the local minima and maxima of the windowed function on $G_{2000}$; an answer \emph{sits on a slope} if it is closer to a ridge than to any floor, and \emph{hedges} if $\min_b|\xhat-u_b|>0.05$.

\paragraph{Uncertainty.} Every spread is read as a Laplace scale $s$ centred on the answer, with $e=\xhat-\xs$ and $a=|e|$. Size: $\bar s/\bar a$. Coverage: $P(a\le ks)$ for $k=1,2,3$, nominal $1-e^{-k}$ ($63.2$, $86.5$ and $95.0\%$); \emph{dev} is the mean absolute deviation of the three from nominal. PIT: $u=F(\xs)$ with $F$ the predictive distribution function, and the PIT error is the mean of $|\hat P(u\le q)-q|$ over $q=0.05,0.10,\dots,0.95$. ENCE: cases are split into ten equal bins by $s$, and ENCE is the mean over bins of $|\bar a-\bar s|/\bar s$ (the Laplace form of the measure of \citet{levi2022evaluating}). NLL: $\log2s+a/s$. CRPS: $s\,(z+e^{-z}-\tfrac34)$ with $z=a/s$, the closed form for a Laplace at distance $a$ from its centre; for sample-based predictives, the sample CRPS. $\rho$: Spearman's rank correlation of $s$ and $a$. Risk--coverage: the cases are sorted by increasing $s$, the risk at coverage $c$ is the mean $a$ over the first $c$ of them, AURC is its area, and E-AURC is AURC minus the AURC of the ordering by $a$ itself. \emph{Huber read-back}: for a Laplace error of scale $b$, $\mathbb E[\mathrm{Huber}_\beta(e)]=\tfrac{b^2}{\beta}\big(1-e^{-\beta/b}\big)$, an increasing function of $b$; the point spread of the \st{} and of the \sttwo, trained to $\mathbb E[\mathrm{Huber}_{0.05}(e)]$, is mapped through its inverse (exact under the Laplace model; nothing fitted; checked on simulated losses). \emph{Mixture readings} of the two-head models: the mixture has weights $w_k$ (softmax), locations $\mu_k$ (sigmoid) and Laplace scales $b_k$ (softplus plus 0.01), $k=1,\dots,5$. The \emph{mode} answer is $\mu_{k^*}$ with $k^*=\arg\max_k w_k$, and its spread is $b_{k^*}$; the \emph{mixture mean} is $\bar x=\sum_k w_k\mu_k$, and its spread is $\mathbb E|X-\bar x|=\sum_k w_k\big(|\mu_k-\bar x|+b_k e^{-|\mu_k-\bar x|/b_k}\big)$, floored at 0.01; the full mixture is scored by its NLL, CRPS and PIT computed numerically. \emph{Recalibration}: the cases of each (family, level) cell are assigned alternately to two halves; for each case set (in-family, hard, out-of-family, heavy-tailed) and each half, $s$ is multiplied by one factor $k=\frac1n\sum_i a_i/s_i$ computed on the other half of that set, the maximum-likelihood Laplace scale factor.

\subsection{The case sets}
\label{app:casesets}

All in-family sets are built by the same routine. For each family in the order of Appendix~\ref{app:families} and each noise level in increasing order, an integer seed is incremented from a base; the global NumPy stream is seeded with it and the training generator draws a function (with acceptance); the stream is seeded again with the same seed and the training sampler draws the twenty locations, bracketing $\xs$ by $\pm0.02$ (up to 400 draws); the noise comes from \texttt{RandomState}(seed $+17$). The case takes the first $N$ seeds per cell that succeed.

\begin{description}\itemsep1pt
\item[In-family (\nScored).] Base 3{,}100{,}000; $\lambda\in\{0,0.25,0.5,1\}$; $N=100$.
\item[Hard (\nHard).] Base 3{,}200{,}000; $\lambda\in\{2,4\}$; $N=40$.
\item[Heavy-tailed (\nHeavy).] Base 3{,}600{,}000; $\lambda\in\{0.25,0.5,1\}$; $N=25$; noise $\lambda d\,t/\sqrt3$ with $t$ Student-$t$ with three degrees of freedom (unit variance). Its Gaussian counterpart in Section~\ref{sec:results-accuracy} is the matching subset of the in-family cases.
\item[Ties (\nTies).] A \texttt{RandomState}(3{,}300{,}000) draws 100 shapes: separation $\Delta\sim U(0.30,0.60)$, $x_1\sim U(0.08,0.92-\Delta)$, $x_2=x_1+\Delta$, barrier $B\sim U(0.5,4)$, $a=16B/\Delta^4$ and a random sign $\pm$. For each gap $g\in\{0,0.02,0.10\}$ the function is $a\big((x-x_1)(x-x_2)\big)^2+t(x-x_1)$ with $t=\pm gB/\Delta$, so that the wells differ by about $gB$; for each noise level $\sigma\in\{0,0.05,0.10,0.20\}\times B$ a case is drawn with seed $3{,}300{,}000+100s+10\,i_g+i_\sigma$, the samples bracketing both wells.
\item[Unseen (\nUnseen) and monitored (\nMonitored).] Bases 3{,}400{,}000 and 3{,}500{,}000; for each function (Table~\ref{tab:oofspec}) and each $\lambda\in\{0,0.25,0.5,1\}$, seeds are incremented one per attempt (at most 400 per cell); an attempt draws, from \texttt{RandomState}(seed), a window of length $L=\omega S$ with $\omega\sim U(0.6,1)$ and $S$ the native domain length, a left end $\sim U(\mathrm{lo},\mathrm{hi}-L)$ and a mirror with probability $\tfrac12$, and is rejected unless the windowed function is finite on $G_{2000}$, $\xs\in(0.05,0.95)$, the best rival basin floor lies at least 2\% of the range above the global minimum, and $d>0$; samples and noise ($\sigma=\lambda d$) follow as above, with $y$ in the function's native units. $N=20$. Rough bowl (unseen) and Griewank (monitored) produced no admissible window, because every window had a rival floor within 2\% of the range, so eleven and five functions are used.
\end{description}

\paragraph{Start-point and path subsets.} The start-point test uses the first 20 in-family cases of every (family, level) cell (1{,}040 cases), each rolled out from the spline argmin and from 0.1, 0.3, 0.7 and 0.9. The solution path is read from each walk model's own trajectory on all in-family cases (and, for regret and hedging along the walk, on all in-family and tie cases).

\paragraph{Griewank (600).}
\label{app:six}
Griewank, $f(x)=1+x^2/4000-\cos x$ on $[-10,10]$, is viewed through a window of its native domain chosen so that the global minimum lands at $\ell\in\{0.18,0.34,0.50,0.66,0.82\}$: with $p=0.5$ the normalised native minimiser, the window width is $W=\min(p/\ell,(1-p)/(1-\ell),1)$ and the case function is $f(u)=f_{\mathrm{native}}\big(-10+20(p-\ell W+uW)\big)/(20W)$, so that $\xs=\ell$, the curve is the published function zoomed without distortion, and $y$ is divided by the factor by which $x$ was compressed. Only the centred window ($\ell=0.5$) contains both valleys next to the global one; each of the other four contains one of them. The neighbouring floors lie 0.49\% of the native range above the global one. Cases are indexed by level $t_i$ (the six levels $\lambda=0$ to $4$), location $l_i$ and repeat $r=0,\dots,19$, with seed $3\cdot10^6+10^5t_i+10^4l_i+r$; for attempt $k=0,\dots,299$ the global stream is seeded with seed $+\,7919k$ and the training sampler is called, the draw being accepted if its points are distinct and bracket $\xs$ by $\pm0.02$; the noise is $\mathcal N(0,(\lambda d)^2)$ from \texttt{RandomState}(seed $+31$). These are the Griewank cases of a six-function test built during development, unchanged.

\paragraph{Noise check (780 $\times$ 3) and Bayes check (900).} For the noise check, seeds run from 9{,}500{,}001, 60 functions per family; each function is scored with $\lambda$ cycling through the six training values, with $\lambda\sim\log U(0.1,4)$, and with $\sigma=cR$, $c\sim\log U(0.003,0.3)$, all three sharing the function and the sample locations. For the Bayes check, seeds run from 9{,}900{,}001 and 300 double-well cases are drawn at each $\lambda\in\{1,2,4\}$; the prior draws use \texttt{RandomState}(20{,}260{,}929).

\begin{table}[t]
\centering\small
\setlength{\tabcolsep}{4pt}
\resizebox{\linewidth}{!}{%
\begin{tabular}{lll}
\toprule
Function & $f(x)$ & Domain \\
\midrule
\multicolumn{3}{l}{\textit{Unseen (never used in development)}} \\
Forrester        & $(6x-2)^2\sin(12x-4)$ & $[0,1]$ \\
Gramacy--Lee     & $\sin(10\pi x)/(2x)+(x-1)^4$ & $[0.5,2.5]$ \\
Ackley           & $-20e^{-0.2|x|}-e^{\cos2\pi x}+20+e$ & $[-4,4]$ \\
Michalewicz      & $-\sin x\,\sin^{20}(x^2/\pi)$ & $[0,\pi]$ \\
Shubert          & $\sum_{j=1}^{5}j\cos\big((j+1)x+j\big)$ & $[-2,4]$ \\
Easom (slice $x_2=\pi$) & $\cos x\;e^{-(x-\pi)^2}$ & $[-1,7.5]$ \\
Drop-wave        & $-\big(1+\cos12|x|\big)/\big(0.5x^2+2\big)$ & $[-2.5,2.5]$ \\
Absolute ripple  & $|x-0.37|+0.3|\sin8x|$ & $[-0.4,1.2]$ \\
Langermann       & $-\sum_{i}c_ie^{-(x-a_i)^2/\pi}\cos\big(\pi(x-a_i)^2\big)$, $(c_i,a_i)=(1,3),(2,5),(5,2),(2,1),(3,7)$ & $[0,10]$ \\
Sinc             & $-\sin(x)/x$ & $[-9,9]$ \\
Damped cosine    & $-e^{-0.35|x|}\cos2.2x+0.02x$ & $[-6,6]$ \\
Rough bowl (no admissible case) & $x^2+0.12\sum_{k=0}^{3}0.5^k\cos(2\pi3^kx)$ & $[-1,1]$ \\
\midrule
\multicolumn{3}{l}{\textit{Monitored (printed during development; not independent)}} \\
Sphere & $x^2$ & $[-5,5]$ \\
Rastrigin & $10+x^2-10\cos2\pi x$ & $[-5.12,5.12]$ \\
Styblinski--Tang & $\tfrac12(x^4-16x^2+5x)$ & $[-5,5]$ \\
Levy & $\sin^2\pi w+(w-1)^2(1+\sin^2 2\pi w)$, $w=1+\tfrac{x-1}4$ & $[-10,10]$ \\
Schwefel & $418.9829-x\sin\sqrt{|x|}$ & $[-500,500]$ \\
Griewank (no admissible case; see the Griewank test) & $1+x^2/4000-\cos x$ & $[-10,10]$ \\
\bottomrule
\end{tabular}}
\caption{The out-of-family functions, in their native form and domain; the Forrester and Gramacy--Lee functions are from \citet{forrester2008engineering} and \citet{gramacy2012cases}, and Ackley, Michalewicz, Shubert and Easom are among the functions catalogued by \citet{jamil2013literature}.}
\label{tab:oofspec}
\end{table}

\paragraph{Note on the unseen set.}
\label{app:oof}
Each of the eleven unseen functions contributes 20 cases at each of the four scored levels (80 cases); the five monitored functions likewise.

\subsection{Classical estimators}
\label{app:estimators}

All surrogate argmins are taken between the first and last sample, where the samples bracket the function; outside it the fits extrapolate.
\begin{description}\itemsep1pt
\item[Lowest observation.] $x_{\arg\min_i y_i}$.
\item[Spline argmin.] The spline of Appendix~\ref{app:inputs}, computed from the raw samples, minimised on $G_{1000}$.
\item[Differential evolution and basin-hopping on the spline.] The same spline; SciPy differential evolution (40 generations, population 12, tolerance $10^{-6}$, polishing) and basin-hopping (12 hops, step 0.25, L-BFGS-B local searches) within $[\min x_i,\max x_i]$, both with seed 0; basin-hopping starts from the lowest observation.
\item[Local quadratic (LOOCV).] Local quadratic regression with a Gaussian kernel, a moving-weight least-squares fit \citep{fan1996local, levin1998approximation}; the bandwidth is chosen from 12 values geometrically spaced in $[0.02,0.5]$ by leave-one-out squared error, and the fit is minimised on 600 points between the first and last sample.
\item[GP.] The observations are standardised by their mean and standard deviation. The kernel is $C\cdot\mathrm{Mat\acute ern}_{5/2}(\ell)+\sigma_n^2$, with $C$ initialised at 1, $\ell$ initialised at 0.15 with bounds $[0.01,1]$, $\sigma_n^2$ initialised at $0.01$ with bounds $[10^{-6},10]$; hyperparameters by maximum marginal likelihood (scikit-learn, two optimiser restarts, seeded by the case). \emph{Posterior argmin}: the minimum of the posterior mean on $G_{1200}$ between the first and last sample. \emph{Argmin posterior}: 256 joint draws of the latent function (the kernel without the white noise) on 256 equally spaced points between the first and last sample, the argmin of each draw jittered uniformly within its grid cell; its \emph{median} is the median of the 256 argmins, its \emph{mode} the median of the draws in the fullest of 40 equal bins, and its \emph{spread} the mean distance of the draws from the reported answer, floored at $0.01$.
\item[Median ensemble.] The median of the \nfm, \st{} and GP posterior-argmin answers, for every pair of \nfm{} and \st{} seeds.
\end{description}

\subsection{Statistics}
\label{app:stats}

Every interval is a 95\% percentile-bootstrap interval of a mean paired difference between two estimators on the same cases. A learned model's value is the mean over its three seeds, and each bootstrap replicate draws three seeds with replacement for each learned model and, independently, resamples cases within each (family, level) cell; the statistic is computed on the drawn seeds and cases. Accuracy and tie measures use 5{,}000 replicates; every uncertainty measure (NLL, CRPS, dev, PIT, ENCE, $\rho$, E-AURC) uses 2{,}000, so its smallest reportable $p$ is $1/2000$. On the tie set the resampling unit is the tie shape (all twelve cases of a shape together), and on the unseen and monitored sets we also resample whole functions. In the Griewank test the five windows are the units. Across families, each family's mean difference is one observation in a Wilcoxon signed-rank test. Each family of declared comparisons (Section~\ref{sec:metrics}, Table~\ref{tab:primary} and Appendix~\ref{app:families-declared}) is adjusted separately: two-sided bootstrap $p$-values (smallest reportable $1/5000$) by Holm's procedure over the comparisons of that family. The walk effect and its change between one and two heads (Table~\ref{tab:walk}) resample cases once per replicate, shared by the four models, and seeds within each model.

\subsection{The Huber loss at a tie}
\label{app:huber}

The position loss of the four learned models' point heads is the smooth-$L_1$ (Huber) loss with threshold $\beta=0.05$, $\ell_\beta(e)=e^2/(2\beta)$ for $|e|<\beta$ and $|e|-\beta/2$ otherwise. Let the label be one of two floors $a<b$ with probability $\tfrac12$ each, a distance $D=b-a\ge2\beta$ apart (in the tie set $D\ge0.30$). For an answer $u$ between the floors and at least $\beta$ from both, the expected loss is $\tfrac12\big[(u-a-\tfrac\beta2)+(b-u-\tfrac\beta2)\big]=\tfrac12(D-\beta)$, the same for every such $u$. At a floor, $u=a$, it is $\tfrac12\big[0+(D-\tfrac\beta2)\big]=\tfrac D2-\tfrac\beta4$, larger by $\beta/4$. Between the two, at $u=a+t$ with $0\le t<\beta$, it is $\tfrac12\big[t^2/(2\beta)+D-t-\tfrac\beta2\big]$, whose derivative $\tfrac12(t/\beta-1)$ is negative, so the expected loss falls monotonically from the floor to the interior value; outside the floors it rises. The answers between the floors and at least $\beta$ from both are therefore exactly the optimal ones, and a floor is worse by $\beta/4$.

\section{The other declared comparisons}
\label{app:families-declared}

Two families of comparisons were declared before the $2\times2$ family of Table~\ref{tab:primary}. Table~\ref{tab:declared} gives both in full, each with its own Holm correction over all its comparisons. Both involve the mixture-only Set Transformer (a Set Transformer whose only output is the five-component mixture, answering with its heaviest component; \paramsMHexact{} parameters, three seeds, trained with the same recipe), which the main text replaces by the \sttwo.

\paragraph{The first declared family.} Eight comparisons against the \nfm{} were declared before the first comparison with the Set Transformers was run. That run used an earlier pair of Set Transformers, with the same architectures, data and recipe but reading $(x_i,y_i,y_i',c_i)$ with $y$ in its own units and four descriptors (\paramsSTearly{} and \paramsMHearly{} parameters). Because it mixed a difference in inputs with a difference in architecture, we then trained the pair on the \nfm's inputs that this paper reports, and scored the same eight comparisons on the same cases. The table gives both runs. The conclusions on location error (tied), regret against the \st{} (lower for the \nfm) and the likelihood against the \st{} (better for the \nfm) are the same in both; with the earlier pair, the \nfm's E-AURC advantage over the mixture-only model survived the correction (Holm $p=0.015$), and with the reported pair it does not ($0.085$). On the same cases the mixture-only model reaches an in-family error of 6.50 and a regret of 7.52 (the earlier one 6.99 and 8.46), and 9.04 on the unseen functions.

\paragraph{The analysis plan.} Before the second and third seeds of the \nfmmh{} were trained, we wrote down five comparisons: the \nfmmh's point answer against the \nfm{} on in-family regret, in-family error and tie hedging, and the \nfmmh's mode against the mixture-only model on in-family regret and error. The first three come from the run of the main text. The two against the mixture-only model are computed from the per-case answers of that run and of the run that scored the mixture-only model, on identical cases (case fingerprint \texttt{e69fc72b0fb9}), with the same bootstrap (cases resampled within cells, seeds within models, 5{,}000 draws). The mixture loss lowers the walk's regret and error, and the \nfmmh's mode has lower regret than the mixture-only model, all significant after the correction; the \nfmmh's mode and the mixture-only model are tied on error, and the \nfmmh's point answer does not hedge significantly less than the \nfm's.

\begin{table}[ht]
\centering\small
\setlength{\tabcolsep}{4pt}
\begin{tabular}{llrrr}
\toprule
Measure & Comparison & Difference [95\% CI] & $p$ & Holm $p$ \\
\midrule
\multicolumn{5}{@{}l}{\textit{First declared family, other $-$ \nfm; reported pair of Set Transformers (\nfm's inputs)}} \\
In-family error & \st & $+0.02$ [$-0.14$, $+0.18$] & 0.80 & 0.80 \\
In-family error & mixture-only ST & $-0.21$ [$-0.47$, $+0.05$] & 0.099 & 0.30 \\
In-family error & GP posterior argmin & $+1.30$ [$+1.00$, $+1.60$] & $<$0.0002 & \textbf{$\le$0.0016} \\
In-family regret & \st & $+2.90$ [$+2.12$, $+3.73$] & $<$0.0002 & \textbf{$\le$0.0016} \\
Tie hedging (points) & \st & $-5.9$ [$-18.1$, $+3.8$] & 0.35 & 0.69 \\
NLL & \st, read-back & $+0.09$ [$+0.05$, $+0.13$] & $<$0.0005 & \textbf{$\le$0.003} \\
E-AURC & \st & $+0.15$ [$+0.03$, $+0.28$] & 0.021 & 0.085 \\
E-AURC & mixture-only ST, derived & $+0.16$ [$+0.03$, $+0.28$] & 0.017 & 0.085 \\
\midrule
\multicolumn{5}{@{}l}{\textit{First declared family, other $-$ \nfm; earlier pair ($y$ in its own units), the declared run}} \\
In-family error & \st & $+0.18$ [$-0.10$, $+0.44$] & 0.21 & 0.61 \\
In-family error & mixture-only ST & $+0.27$ [$-0.19$, $+0.70$] & 0.22 & 0.61 \\
In-family error & GP posterior argmin & $+1.30$ [$+1.00$, $+1.60$] & $<$0.0002 & \textbf{$\le$0.0016} \\
In-family regret & \st & $+2.10$ [$+1.21$, $+2.93$] & $<$0.0002 & \textbf{$\le$0.0016} \\
Tie hedging (points) & \st & $-2.0$ [$-14.4$, $+8.8$] & 0.76 & 0.76 \\
NLL & \st, read-back & $+0.10$ [$+0.04$, $+0.16$] & $<$0.0005 & \textbf{$\le$0.003} \\
E-AURC & \st & $+0.16$ [$+0.02$, $+0.30$] & 0.028 & 0.11 \\
E-AURC & mixture-only ST, derived & $+0.31$ [$+0.13$, $+0.48$] & 0.003 & \textbf{0.015} \\
\midrule
\multicolumn{5}{@{}l}{\textit{Analysis plan, \nfmmh{} $-$ other}} \\
In-family regret & \nfmmh, point $-$ \nfm & $-2.71$ [$-3.33$, $-2.08$] & $<$0.0002 & \textbf{$\le$0.001} \\
In-family error & \nfmmh, point $-$ \nfm & $-0.44$ [$-0.64$, $-0.25$] & $<$0.0002 & \textbf{$\le$0.001} \\
In-family regret & \nfmmh, mode $-$ mixture-only ST & $-0.99$ [$-1.53$, $-0.47$] & 0.0004 & \textbf{0.0012} \\
In-family error & \nfmmh, mode $-$ mixture-only ST & $-0.15$ [$-0.44$, $+0.14$] & 0.30 & 0.60 \\
Tie hedging (points) & \nfmmh, point $-$ \nfm & $-6.0$ [$-18.8$, $+5.6$] & 0.37 & 0.60 \\
\bottomrule
\end{tabular}
\caption{The two earlier declared families, in full, each with its own Holm correction (over eight and over five comparisons). Bold: significant after the correction. In the first family positive values favour the \nfm{} (for tie hedging, a positive value means the other model hedges more); in the analysis plan negative values favour the \nfmmh. $<$ and $\le$ mark $p$-values at the floor of the bootstrap (Table~\ref{tab:primary}). The reported pair's rows come from a run of the same harness on the same cases with the \nfm, the two Set Transformers of the pair and the classical estimators; the \nfm's and the classical estimators' numbers there are those of the main text.}
\label{tab:declared}
\end{table}

\begin{table}[ht]
\centering\small
\setlength{\tabcolsep}{4pt}
\begin{tabular}{lrrrr}
\toprule
& \multicolumn{4}{c}{Noise $\sigma/B$} \\
\cmidrule(lr){2-5}
Answer & 0 & 0.05 & 0.1 & 0.2 \\
\midrule
\nfm{} & 15.7 & 26.3 & 31.3 & 27.7 \\
\nfmmh, point & 11.0 & 19.3 & 24.0 & 24.3 \\
\st{} & 13.0 & 15.3 & 20.0 & 24.0 \\
\sttwo, point & 10.7 & 11.7 & 16.3 & 20.3 \\
\nfmmh, mode & 0.0 & 0.0 & 0.0 & 0.3 \\
\sttwo, mode & 0.0 & 0.0 & 0.0 & 0.0 \\
\nfmmh, mixture mean & 53.3 & 63.7 & 78.3 & 87.7 \\
\sttwo, mixture mean & 42.7 & 59.0 & 64.0 & 80.7 \\
GP posterior argmin & 0.0 & 0.0 & 1.0 & 2.0 \\
GP argmin posterior, mode & 0.0 & 1.0 & 2.0 & 5.0 \\
\bottomrule
\end{tabular}
\caption{Hedging (\%) on the tie set where the wells differ by 10\% of the barrier, by noise level (100 shapes per cell; learned models, three seeds each). With noiseless samples the deeper well is identifiable, yet the point answers hedge in 11 to 16\% of cases.}
\label{tab:hedgenoise}
\end{table}

\section{Further results}
\label{app:further}

\subsection{The walk: start points}
\label{sec:results-path}

Whether a walk's answer depends on where it starts was tested on 1{,}040 in-family cases (20 from every family and level), each solved from five starting points (the spline argmin, 0.1, 0.3, 0.7 and 0.9). The five endpoints of a case lie within a median of 0.50 to 0.55\% of the domain of each other for the \nfm{} and 0.37 to 0.45\% for the \nfmmh{} (by seed), they end in two or more different valleys in 2.1 to 5.1\% and 3.4 to 4.5\% of cases, and the mean error by starting point varies by at most 0.2 points. Over the last five steps, 57\% (\nfm) and 41\% (\nfmmh) of the answers move by less than 0.5\% of the domain. In practice the walk behaves as a recurrent refinement rather than a search; every intermediate position is available for inspection, a property neither Set Transformer has.

\subsection{Checks on the comparison}
\label{sec:results-checks}

\paragraph{Is the noise model a shortcut?} Training and every evaluation set tie the noise to the depth of the global well, $\sigma=\lambda d$, with $\lambda$ from six fixed values. A model that estimates $\sigma$ could therefore learn something about $d$ that no real problem would reveal. We built 780 new cases from the training families (60 per family) and scored them three times with identical functions and sample locations: with $\lambda$ from the six training values; with $\lambda$ drawn log-uniformly from $[0.1,4]$; and with $\sigma$ tied to the function's range instead of its well depth, $\sigma=cR$ with $c$ log-uniform in $[0.003,0.3]$, so that $\sigma$ carries no information about $d$. The ordering does not change. On the training values the \nfm, \nfmmh, \st{} and \sttwo{} reach 9.42, 9.14, 9.67 and 9.28, against 11.63 for the GP; with continuous $\lambda$, 9.01, 8.57, 9.15, 8.78 and 11.25; with range-tied noise, 7.50, 7.10, 7.69, 7.45 and 9.00 (\nfm{} $-$ GP $-1.50$ [$-2.32$, $-0.67$], \nfm{} $-$ \st{} $-0.20$ [$-0.64$, $+0.27$]). Grouped by the effective level $\sigma/d$, the \nfm's advantage over the GP under range-tied noise is $-2.9$, $-1.9$, $-3.4$ and $-3.5$ points in the bins from 0.35 upward (significant in three of the four), and absent below 0.35 ($-0.01$), as on the training levels. We find no evidence that the models exploit the tie between noise and depth.

\paragraph{Stronger classical baselines.} Two objections to the GP posterior argmin as the classical reference are that it answers with the minimum of the posterior mean rather than with a summary of the posterior distribution of the minimiser, and that a smoother tuned for the location would do better. The harness therefore scores the median and the mode of the GP's argmin posterior and a local quadratic regression with a cross-validated bandwidth. The posterior median is the strongest classical estimator on location error (7.79 in family) but remains behind the \nfm{} by 1.08 [0.79, 1.37] points; the mode reaches 9.04 and the local quadratic 11.10.

\paragraph{A Bayes basin choice on one family.} That learned models reach similar right-basin rates does not show that the rate is the limit of what the samples contain. For the double-well family, whose five parameters are drawn from a known prior (Appendix~\ref{app:families}), the limit can be computed. We drew four million functions from the family's prior, weighted each by the likelihood of a case's twenty noisy samples (Gaussian noise at each of the six levels with equal prior weight, and a free vertical offset, integrated out under a flat prior, since the inputs are shifted by their minimum), and took as the Bayes answer the point with the largest posterior probability of lying in the global basin of the true function, the rule that maximises the right-basin rate (each prior draw contributes its weight to every point of its own global basin). On 300 new double-well cases per level, this rule chooses the right basin in 81.0\% of cases at $\lambda=2$, against 68.6\% for the \nfm, 70.4\% for the \nfmmh, 73.4\% for the \st{} and 74.1\% for the \sttwo{} (rule $-$ \nfm: $+12.4$ [$+6.0$, $+18.9$] points; rule $-$ \st: $+7.6$ [$+1.3$, $+13.8$]), and in 70.7\% at $\lambda=4$, against 67.1, 69.0, 69.9 and 70.9\% (rule $-$ \nfm: $+3.6$ [$-3.6$, $+10.9$]). Because it is an actual decision rule evaluated on held-out cases, its rate is attainable from the samples, by an estimator that knows this family's prior and targets the basin directly. The learned models do neither: they were trained on thirteen families and on the distance to the minimiser. At $\lambda=2$ the gap therefore shows that a better basin choice is possible on this family, the walk models further from it than the single-pass models; it does not show that the learned models fail to use information that their training objective rewards. At $\lambda=4$ the learned models are within about four points of the rule, and the gap is not significant for the \nfm{} or the \st; the \sttwo's 70.9\% is even slightly above the rule's 70.7\%, which shows that the rule is an approximation there. (Taking instead the mode of the posterior distribution of the minimiser's location, a rule that does not target the basin, gives 73.7 and 64.7\%.) The weights are concentrated: the median effective sample size is 42 at $\lambda=2$ and 215 at $\lambda=4$, but a tenth of the cases rest on about two or fewer effective draws, and the posterior's own estimate of its right-basin rate exceeds the realised one, so the rule is an approximation of the Bayes-optimal one and the true ceiling is at least as high. At $\lambda\le1$ the likelihood is too concentrated for four million prior draws (median effective sample size 7.5 at $\lambda=1$), and we do not report a rate there. This is one family of thirteen, and one where the two wells are far apart.

\subsection{Complementary errors and the ensemble}
\label{sec:results-ensemble}

The median of the \nfm, \st{} and GP posterior-argmin answers, which costs no extra model call, reaches 6.16 on the in-family cases, better than the \nfm{} by 0.56 [0.44, 0.67], than the \st{} by 0.58 and than the GP by 1.86 (all significant), and about as accurate as the two-head models' point answers (6.27 and 6.28). On the hard levels it is level with the best one-head model (13.42 against the \st's 13.41), and on the unseen functions it reaches 8.50, better than the four learned models of the $2\times2$ design (the looped Set Transformer of Section~\ref{sec:results-looped}, scored later, reaches 8.21) but behind the GP posterior argmin (7.85; $+0.65$ [$+0.04$, $+1.29$]) and the GP posterior median (7.55). It was defined before the two-head models existed and is not among the declared comparisons.

\subsection{Griewank}
\label{app:griewank}

\begin{figure}[t]
\centering
\includegraphics[width=\linewidth]{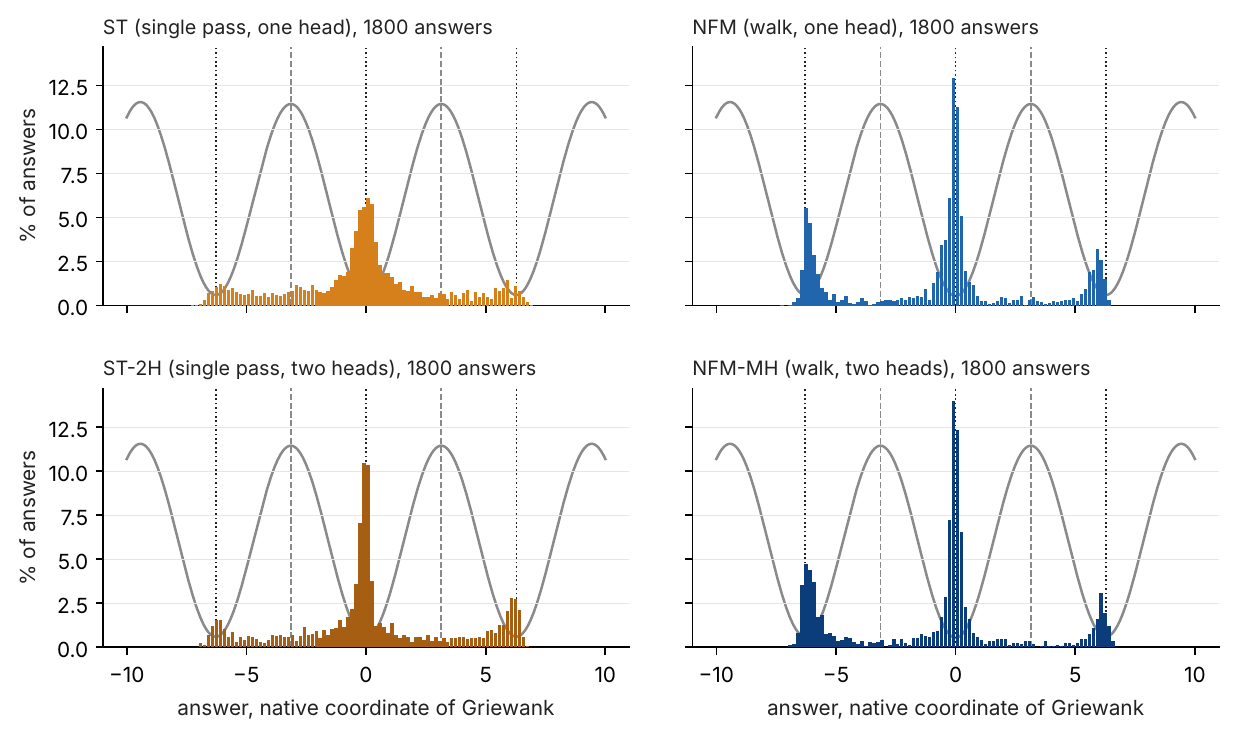}
\caption{Where the point answers land on the tied minima of Griewank: all 600 cases (five windows $\times$ six noise levels $\times$ 20 repeats) and all three seeds of each model, mapped back to the native coordinate, against the function (grey). Dotted lines mark valley floors (the global one at 0 and the two nearly equal ones at $\pm2\pi$), dashed lines the ridges between them. Left, single pass; right, walk; top, one head; bottom, two heads. The single-pass answers spread more across the slopes between the valleys; the walk's concentrate more on the floors.}
\label{fig:griewank}
\end{figure}

The observation has limits: it is measured on synthetic double wells, not on real data. On one function with adjacent valleys the walk and the single pass do differ: on Griewank, viewed through five windows that place its global minimum at five locations, each window containing at least one neighbouring valley whose floor is about 0.5\% of the range above the global one (both neighbours only in the centred window; 600 cases, every seed; Figure~\ref{fig:griewank}), the walk's answers lie on a slope between valleys less often than the single pass's, with one head (\nfm{} 11.4\%, seeds 10.7, 12.8 and 10.7; \st{} 27.1\%, seeds 22.0, 28.7 and 30.7) and with two (\nfmmh{} 9.6\%, seeds 9.2, 8.0 and 11.5; \sttwo{} 21.3\%, seeds 20.3, 19.3 and 24.2), and its regret there is lower (14.0 against 29.7\% of the range with one head, 11.9 against 23.7\% with two; \nfmmh{} $-$ \sttwo{} $-11.8$ [$-16.2$, $-7.3$] points, slope $-11.7$ [$-16.4$, $-6.6$]). Two cautions apply. Griewank was among the functions watched during development, and the effect was first noticed on it, so it is not an independent test; and with only five windows, the interval obtained by resampling them is a coarse guide. The GP answers on a floor in every case. A walk whose steps shrink may end near a floor more often than a regression does when the valleys are close, which agrees with the lower regret it earns in the second half of the walk (Section~\ref{sec:results-walk}), but on the constructed ties, whose valleys are farther apart, the walk hedges as often as the single pass.

\subsection{Inference-time ablations}
\label{app:ablations}

The training monitor measures, at step \trainEpochs{} and on its 480 fixed in-family cases, the effect of replacing one quantity by its average over cases or by another case's (\nfm, means over the three seeds; every change is measured against the intact model on the same 480 cases). Freezing the updater's correction at its case average raises the error by 8.8 points (8.7, 11.4 and 6.4 in the three seeds; 6.6, 8.9 and 6.3 for the \nfmmh); giving every case another case's state $\mathbf e$ raises it by 6.6, another case's descriptors $\mathbf g$ by 3.5, and permuting the observations among cases, the sanity check, by 18.2. Two inputs do almost nothing: replacing the resolution radius $\delta$ fed to the local read with another case's changes the error by less than 0.01 points, and the bandwidth multiplier sits at its lower bound in 72\% of cases; with the multiplier there, the window's width is mostly at its floor of one sample spacing (Section~\ref{sec:model}), which is why $\delta$ changes nothing. The summary vector and the updater carry the answer; the learned bandwidth is unused, and a reduced model without it is the obvious next experiment.

\subsection{Uncertainty: paired differences}
\label{app:calib-diff}

\begin{table}[H]
\centering
{\footnotesize\setlength{\tabcolsep}{3pt}
\begin{tabularx}{\linewidth}{@{}l>{\raggedright\arraybackslash}X@{}}
\toprule
\multicolumn{2}{@{}l}{\textit{Paired differences, walk $-$ single pass (negative NLL, dev, PIT, ENCE, E-AURC and positive $\rho$ favour the walk)}} \\
\midrule
One head: \nfm{} $-$ \st{} (read-back) & NLL $-0.09$ [$-0.13$, $-0.05$]$^{*}$, dev $+0.004$ [$-0.008$, $+0.014$], PIT $+0.001$, ENCE $+0.010$; $\rho$ $+0.059$$^{*}$, E-AURC $-0.154$ [$-0.283$, $-0.009$]$^{*}$ \\
Two heads, point: \nfmmh{} $-$ \sttwo{} (read-back) & NLL $-0.12$ [$-0.19$, $-0.06$]$^{*}$, dev $-0.022$ [$-0.032$, $-0.010$]$^{*}$, PIT $-0.007$, ENCE $-0.068$$^{*}$; $\rho$ $+0.047$$^{*}$, E-AURC $-0.09$ [$-0.23$, $+0.06$] \\
Two heads, mixture mean & NLL $-0.06$$^{*}$, dev $+0.006$, PIT $+0.011$$^{*}$, ENCE $+0.038$; $\rho$ $+0.034$$^{*}$, E-AURC $-0.08$ [$-0.17$, $+0.02$] \\
Two heads, full mixture & NLL $-0.17$ [$-0.21$, $-0.12$]$^{*}$, CRPS $-0.02$ [$-0.14$, $+0.12$], PIT $+0.008$ \\
\bottomrule
\end{tabularx}}
\caption{Paired differences in the uncertainty measures of Table~\ref{tab:calibration}, walk minus single pass, on the \nScored{} in-family cases. Intervals as in Table~\ref{tab:benchmark}, 2{,}000 draws; $^{*}$ marks an interval that excludes zero.}
\label{tab:calib-diff}
\end{table}

\clearpage
\section{Per-seed results}
\label{app:seeds}

Table~\ref{tab:seeds} gives every seed of the four models of the $2\times2$ design.

\begin{table}[H]
\centering\footnotesize
\setlength{\tabcolsep}{4pt}
\begin{tabular}{llrrrrr}
\toprule
Model & Seed & Error & Median error & Regret & Median regret & Right basin \\
\midrule
\nfm & 1 & 6.67 & 2.25 & 12.98 & 2.10 & 82.7\% \\
     & 2 & 6.68 & 2.30 & 12.45 & 2.11 & 82.5\% \\
     & 3 & 6.80 & 2.40 & 12.97 & 2.49 & 82.2\% \\
\st  & 1 & 6.75 & 2.83 & 15.38 & 3.02 & 82.7\% \\
     & 2 & 6.71 & 2.75 & 15.21 & 2.92 & 82.7\% \\
     & 3 & 6.75 & 2.94 & 16.51 & 3.33 & 82.9\% \\
\nfmmh, point & 1 & 6.31 & 1.81 & 10.52 & 1.51 & 82.9\% \\
     & 2 & 6.36 & 1.72 & 10.16 & 1.34 & 82.5\% \\
     & 3 & 6.15 & 1.61 & 9.58 & 1.19 & 83.4\% \\
\sttwo, point & 1 & 6.29 & 2.08 & 13.08 & 1.98 & 83.7\% \\
     & 2 & 6.28 & 2.03 & 13.00 & 1.85 & 82.9\% \\
     & 3 & 6.28 & 2.15 & 13.55 & 2.09 & 83.7\% \\
\nfmmh, mode & 1 & 6.36 & 1.08 & 6.52 & 0.63 & 82.5\% \\
     & 2 & 6.48 & 1.10 & 6.91 & 0.62 & 81.8\% \\
     & 3 & 6.23 & 1.00 & 6.16 & 0.62 & 83.4\% \\
\sttwo, mode & 1 & 6.25 & 1.43 & 7.96 & 1.16 & 83.2\% \\
     & 2 & 6.44 & 1.45 & 7.84 & 1.24 & 82.7\% \\
     & 3 & 6.40 & 1.44 & 8.15 & 1.17 & 82.7\% \\
\bottomrule
\end{tabular}
\caption{Every seed on the \nScored{} in-family cases (error and median error in percent of the domain, regret and median regret in percent of the range). The means over seeds are those of Table~\ref{tab:benchmark}; the median columns there are means of the seeds' medians. Every \nfm{} seed has lower regret than every \st{} seed, and every \nfmmh{} seed than every \sttwo{} seed, for the point answers and for the modes. Computed after all seeds had been scored.}
\label{tab:seeds}
\end{table}

\clearpage
\section{Hyperparameters}
\label{app:hyper}

Table~\ref{tab:hyper} lists the settings of the \nfm{} and its training.

\begin{table}[H]
\centering\small
\begin{tabularx}{\linewidth}{@{}p{0.34\linewidth}X@{}}
\toprule
Samples per case $n$ & 20 \\
Iterations $T$ & 15 \\
Model width $d_{\text{model}}$ / summary width $d_{\mathbf e}$ / iterator hidden & 128 / 64 / 256 \\
Inputs & $y$ normalised by its range $D_y$; five descriptors, the fifth $\mathrm{clip}_{[-8,8]}(\ln D_y)$ \\
Local read & Cauchy window at five dyadic widths $2^{s}h$, gated to $5\cdot2^{s}$ nearest samples; bandwidth multiplier in $[0.5,4]$, floor $1/20$; mixture exponent $\zeta\in[0.25,4]$ \\
Updater & query $98\to256$, tokens $131\to256$, 4 heads, output projection and layer norm, two-layer network of width 496, read-out to 64; residual $\mathbf e+\tanh(\gamma)\,\Delta\mathbf e\,\sqrt{64}/\overline{\lVert\Delta\mathbf e\rVert}$, $\gamma$ initialised at 0, running norm with momentum 0.999, gate gradient divided by 64 \\
Activation & learned clamped cubic, leak 0.02 below zero; coefficients capped at $a_1\le e^{2}$, $a_2\le1$, $a_3\le e^{-2}$ \\
Travel budget / step floor & 2.0 domains / $10^{-4}$ \\
Position loss & Huber, $\beta=0.05$, divided by the constant $0.07$; trajectory weight $0.5$, linearly time-weighted \\
Calibration loss & $|e|/s+\log s$ on the detached error, weight $1.0$ \\
Auxiliary noise loss / out-of-bounds penalty & $0.05$ / $1.0$ \\
Mixture head (\nfmmh{} and \sttwo) & one linear layer on the final state $\mathbf e$ (64) or on the decoder output $H$ (228) to $3\times5$; weights softmax, locations sigmoid, scales softplus $+\,0.01$; biases initialised to weights 0, locations $\mathrm{logit}((k-\frac12)/5)$ for $k=1,\dots,5$, scales $-1$; mixture NLL weight 1.0, gradient to the whole network \\
Unresolvable-case weight & $0.3$ (position and mixture losses) \\
Optimiser & Adam, lr $2\times10^{-4}$, batch 48, gradient clipping at 1.0; cosine schedule over $3\times10^{6}$ steps to $10^{-6}$, stopped at \trainEpochs{} \\
Gradient guard & skip an update whose gradient norm exceeds $100\times$ the running median of accepted norms \\
Start of the rollout & spline argmin (50\%) or uniform random (50\%) \\
Noise levels $\lambda$ & $\{0, 0.25, 0.5, 1, 2, 4\}$, uniform \\
Seeds & drawn at random and recorded in every checkpoint \\
\bottomrule
\end{tabularx}
\caption{\nfm{} and \nfmmh{} training and architecture settings, read from the released training sources (\texttt{train\_source.py} in each run folder), and the mixture head shared with the \sttwo.}
\label{tab:hyper}
\end{table}

\clearpage
\section{Harness self-checks}
\label{app:harness}

Before scoring, the harness (\texttt{NFM\_tests\_two\_heads\_standard.py}; its quick version differs only in its case counts) checks every model folder: the checkpoint at exactly step \trainEpochs, the hashes of the checkpoint and of its training source against the recorded ones, the parameter count, and, for every Set Transformer, that its inputs equal the \nfm's on 200 random curves; a model that fails stops the run, and a seed without the step-\trainEpochs{} checkpoint is skipped and listed. The benchmark then runs the following checks, all of which passed on the reported run: the closed-form Laplace CRPS agrees with numerical integration; all training sources use the same thirteen families and weights (the family tag is compared); on a simulated calibrated model the calibration measures return coverage deviation and PIT error near 0 and $\bar s/\overline{|e|}$ near 1; the Huber read-back recovers the Laplace scale from simulated Huber losses; the sample-based CRPS equals the closed form, the E-AURC of an oracle ordering is 0, and the ENCE of a calibrated model is near 0; the local quadratic regression and the GP argmin posterior find the minimum of a parabola; and changing the batch size changes no answer of any model by more than 0.1\% of the domain. Each run records its library versions, a 12-character hash of every training source and checkpoint, and a fingerprint of the case set (\texttt{e69fc72b0fb9} for the standard run).

The standard run reported here took 48 minutes for the benchmark and 2.3 hours in all on a two-core machine (Python 3.11, NumPy 2.4, PyTorch 2.14), and was repeated in full on a two-core Colab CPU (Python 3.13, NumPy 2.1, scikit-learn 1.6, PyTorch 2.11). The case fingerprints were identical. Of 310{,}240 per-case answers, those of the two Set Transformers agree to within $10^{-5}$ and those of the \nfmmh{} to within $2.3\times10^{-4}$ of the domain; the \nfm's differ by more than $10^{-4}$ on 18 cases (at most 0.4\% of the domain), from floating-point differences between library versions, and the stochastic classical estimators (the GP's argmin draws and basin-hopping) differ on 15 cases. Every printed number of the benchmark, the Griewank test, the noise and Bayes checks and the ensemble is identical in the two runs or differs by one unit in its last printed digit (at most 0.01 in the learned models' results). Every number in this paper is taken from the reported machine.

The Griewank tie test (Section~\ref{sec:results-tie}) uses the case construction of Appendix~\ref{app:six}; it loads every model from its own training source and builds each model's inputs with the input builder recorded with it (for the Set Transformers, the \nfm's), refuses two folders that hold the same weights, and accepts only the step-\trainEpochs{} checkpoint. The moving-viewpoint probe of Section~\ref{sec:results-ablations} is the \texttt{probe} section of the same script (run separately, 7 minutes on two cores); it first checks that its own rollout reproduces each model's forward pass exactly. The noise and Bayes checks of Appendix~\ref{sec:results-checks} reuse the case code of a reproduction script written during development and give each model its own inputs in the same way; their cases are generated by the shared data source of the training pipeline, the same functions that generated the training data.

\end{document}